\documentclass[3p]{elsarticle}
\usepackage{cellspace} 
\usepackage{doi}      

\usepackage[font=normalsize]{caption}
\usepackage{booktabs}
\usepackage{siunitx} 
\usepackage{makecell}
\usepackage{float}
\usepackage{graphicx}
\usepackage{float}
\usepackage{algorithm}
\usepackage{amsmath}
\usepackage{algpseudocode}
\usepackage{rotating}

\usepackage{hyperref}

\usepackage{amssymb}
\usepackage{amsmath}

\usepackage{longtable}
\usepackage{array}
\usepackage{booktabs} 
\usepackage{geometry} 
\usepackage{caption}  

\usepackage{soul} 
\usepackage{xcolor} 

\usepackage{tcolorbox}

\usepackage{mdframed}
\usepackage{xcolor}
\usepackage{caption}   
\usepackage{float}     
\usepackage{geometry}  
\usepackage{lipsum}    

\usepackage{afterpage}

\usepackage{soul} 
\soulregister{\cite}{7} 
\soulregister{\ref}{7}
\soulregister{\label}{7}

\journal{Energy and Buildings}

\begin{document}

\begin{frontmatter}



\title{Explaining Deep Learning-based Anomaly Detection in Energy Consumption Data by Focusing on Contextually Relevant Data}


\author[1]{Mohammad Noorchenarboo}
\author[1]{Katarina Grolinger\corref{cor1}}

\affiliation[1]{organization={Department of Electrical and Computer Engineering, Western University}, 
            addressline={1151 Richmond St}, 
            city={London},
            postcode={N6A 3K7}, 
            state={Ontario},
            country={Canada}}

\cortext[cor1]{Corresponding author}
\emailauthor{Katarina Grolinger}{kgroling@uwo.ca}

\begin{abstract}
Detecting anomalies in energy consumption data is crucial for identifying energy waste, equipment malfunction, and overall, for ensuring efficient energy management. Machine learning, and specifically deep learning approaches, have been greatly successful in anomaly detection; however, they are black-box approaches that do not provide transparency or explanations. SHAP and its variants have been proposed to explain these models, but they suffer from high computational complexity (SHAP)  or instability and inconsistency (e.g., Kernel SHAP). To address these challenges, this paper proposes an explainability approach for anomalies in energy consumption data that focuses on context-relevant information. 
The proposed approach leverages existing explainability techniques, focusing on SHAP variants, together with global feature importance and weighted cosine similarity to select background dataset based on the context of each anomaly point. By focusing on the context and most relevant features, this approach mitigates the instability of explainability algorithms.
Experimental results  across 10 different machine learning models, five datasets, and five XAI techniques, demonstrate that our method reduces the variability of explanations providing consistent explanations. Statistical analyses confirm the robustness of our approach, showing an average reduction in variability of approximately 38\% across multiple datasets.

\end{abstract}

\begin{graphicalabstract}
\scalebox{0.5}{\includegraphics{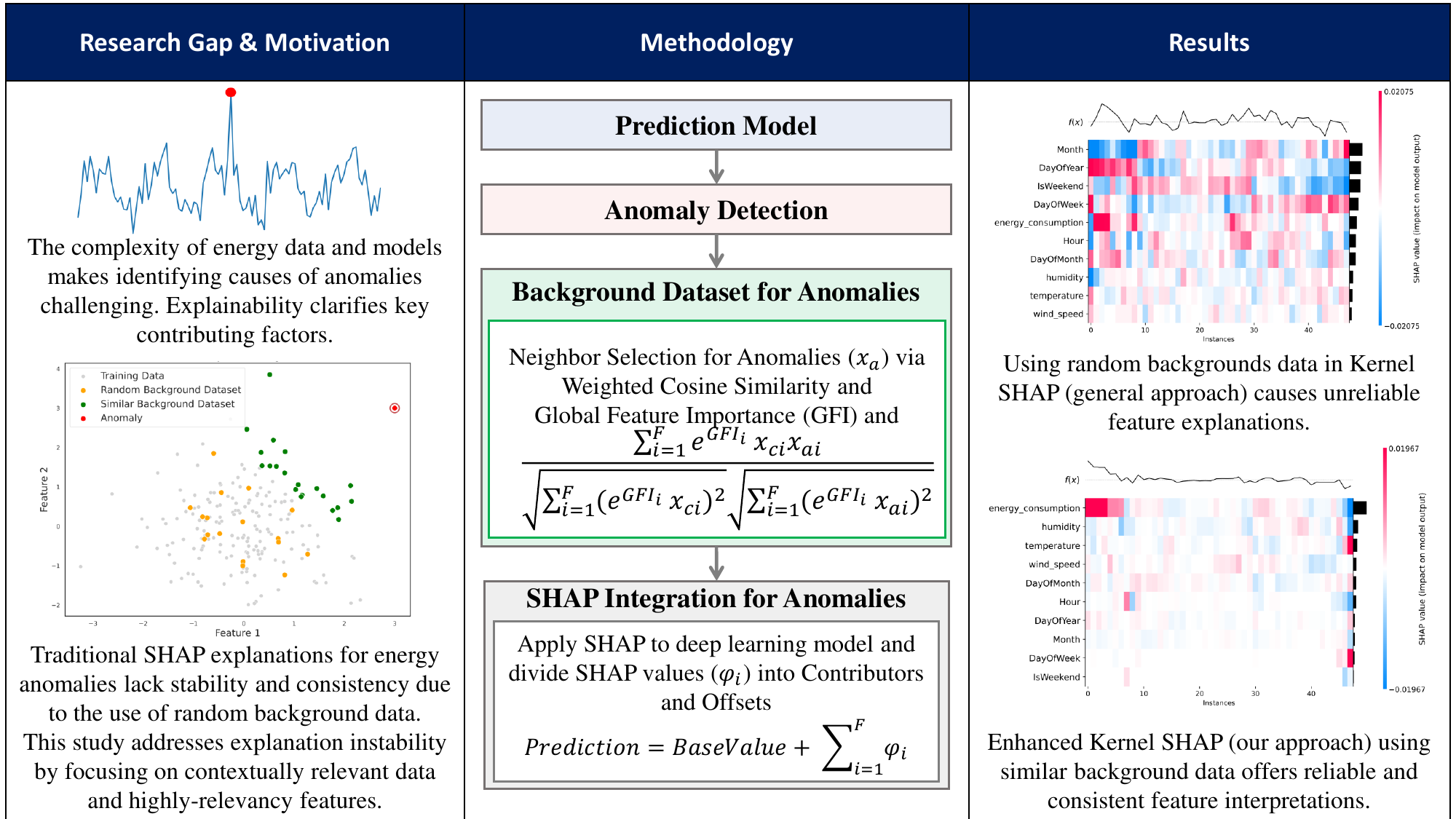}}
\end{graphicalabstract}


\begin{highlights}
\item SHAP-based explanations are hindered by SHAP’s computational complexity 
\item SHAP variants improve this but suffer from instability
\item We propose context-focused explanations for energy consumption data
\item Our approach reduces SHAP variability by approximately 38\%
\item Statistical analyses confirm the robustness of our explainability approach

\end{highlights}


\end{frontmatter}

\begin{table}[H]
\caption*{List of Acronyms and Symbols}
 \begin{tabular}{l|l} 
    \toprule
    Acronym/Symbol & Definition\\
    \midrule
    1D-CNN & One-dimensional Convolutional Neural Network\\
AI & Artificial Intelligence\\
\( \mathrm{Base Value} \) & Average prediction made using the background dataset\\
BGRU & Bidirectional Gated Recurrent Unit\\
BLSTM & Bidirectional Long Short-Term Memory\\
CNN & Convolutional Neural Network\\
DCNN & Dilated Convolutional Neural Network\\
DL & Deep Learning\\
DNN & Deep Neural Network\\
\( e \) & Prediction error\\
\( |F| \) & Total number of features in the model\\
\( GFI_i \) & Global Feature Importance score for feature \( i \)\\
\( GFI_i' \) & Exponentially transformed Global Feature Importance score for feature \( i \)\\
GRU & Gated Recurrent Unit\\
\( h \) & Forecasting horizon\\
\( I \) & Number of time steps for inputs sequence\\
\( IQR \) & Interquartile range, calculated as \( Q_3 - Q_1 \)\\
\( k(F, S) \) & Kernel weights ensuring fair evaluation of feature contributions\\
LIME & Local Interpretable Model-Agnostic Explanations\\
LSTM & Long Short-Term Memory\\
\( O_i \) & Output sequence of target values in horizon $i$\\
\( Q_1 \) & First quartile (25th percentile) of the prediction errors\\
\( Q_3 \) & Third quartile (75th percentile) of the prediction errors\\
RNN & Recurrent Neural Network\\
\( S(x_c, x_a) \) & Weighted cosine similarity score between \( x_c \) and \( x_a \)\\
SD & Standard Deviation\\
\( |S| \) & Number of features in subset \( S \)\\
SHAP & Shapley Additive Explanations\\
TCN & Temporal Convolutional Network\\
TFT & Temporal Fusion Transformer\\
TPE & Tree-structured Parzen Estimator\\
TST & Time Series Transformer\\
\( v(S) \) & Model value function evaluated for subset \( S \)\\
WLS & Weighted Least Squares\\
XAI & Explainable Artificial Intelligence\\
\( X \) & Original feature values\\
\( X' \) & Scaled feature values\\
\( X_{f,t} \) & Value of feature \( f \) at time \( t \)\\
\( X_{\mathrm{max}} \) & Maximum value of the feature\\
\( X_{\mathrm{min}} \) & Minimum value of the feature\\
\( x_{ai} \) & Value of feature \( i \) for the anomalous sample\\
\( x_{ci} \) & Value of feature \( i \) for a sample from the training dataset\\
\( Y_{t+h} \) & Target energy consumption values for time steps \( t+h \)\\
\( \binom{|F|}{|S|} \) & Binomial coefficient representing the number of ways to choose \( |S| \) features from \( |F| \)\\
\( \phi_0 \) & Base value representing the average prediction over the entire background dataset\\
\( \phi_i \) & SHAP value representing feature \( i \)'s contribution to the prediction\\
   
    
    \bottomrule
 
 \end{tabular}
\end{table}

\section{Introduction}
\label{sec:introduction}


The rising global demand for energy, particularly electricity, is causing significant environmental impacts, including increased greenhouse gas emissions and the depletion of vital resources \cite{fan2018analytical}. Global electricity demand is anticipated to rise by nearly 80\% by 2040 \cite{martin2018optimized}. Residential and commercial buildings, accounting for one-third of global energy consumption, are major contributors to these environmental impacts \cite{fan2018analytical}. Improving building energy efficiency, especially in terms of electricity usage, is essential to mitigate the adverse effects of growing energy consumption \cite{pan2022high}. Achieving this goal relies heavily on detecting and correcting anomalies in electricity consumption. These anomalies, defined as irregularities or deviations from normal energy behavior, include unusual consumption patterns caused by faulty device operations, user negligence (e.g., leaving windows or refrigerator doors open), theft, or non-technical losses. If not addressed promptly, these issues can result in energy waste, increased power consumption, and devices running longer than necessary due to inefficiencies or malfunctions, leading to additional energy waste and potential equipment damage \cite{dai2021smart,liu2021data,himeur2021artificial}.


Data-driven approaches have proven effective in identifying anomalies, offering reliable alerts to analysts and energy managers \cite{xu2020hybrid,10620263}. Anomaly detection has evolved from traditional statistical methods, which struggle with complex structures and large datasets, to Deep Learning (DL) methods that automatically learn from time series data \cite{li2023deep, Esen_2009_ANN_wavelet, Esen_2015_experimental}. DL algorithms excel in identifying abnormal electricity consumption patterns due to their ability to model complex non-linear relationships and leverage multi-layered architectures for hierarchical feature extraction \cite{qu2021combined, Esen_2008_ANN_neurofuzzy}. Despite their superior accuracy, DL models face challenges in transparency and explainability, crucial factors for building trust and ensuring successful real-world deployment \cite{yepmo2022anomaly}. Providing clear, instance-specific explanations for anomalies is needed to enhance expert trust, support informed decisions, and facilitate the adoption of complex DL models in the energy sector \cite{Antwarg2021}.


Explainable Artificial Intelligence (XAI) aims to enhance transparency and provide explanations, allowing users to understand and trust them \cite{ravi2021general}. While XAI methods have been primarily used in computer vision to provide visual interpretability and explain decisions in tasks such as object recognition, they are equally important for time series data to understand decision-making processes. As sensors become more affordable and ubiquitous, generating vast amounts of electricity consumption time series data, analysis of these data can automate tasks such as energy usage monitoring to enhance maintenance and reduce inefficiencies. For DL-based time series models, such as Long Short-Term Memory (LSTM) networks or Transformers, data are typically transformed into time-series segments using the sliding window technique, necessitating the adaptation of current explainability approaches to provide meaningful explanations for electricity consumption analysis based on this segmented data structure. Converting time points into features for local feature importance and visualizing relevance, similar to saliency masks in images, will provide experts with insights into the decision-making process and facilitate energy improvement tasks \cite{schlegel2019towards}.


While XAI approaches are generally categorized as either model-specific, tailored to particular model architectures, or model-agnostic, applicable to any model by focusing on input-output relationships, this study focuses on model-agnostic methods due to their versatility and broad applicability across diverse machine learning models. A prominent model-agnostic approach is SHapley Additive exPlanations (SHAP), which is widely recognized for its ability to provide consistent and interpretable feature importance scores using Shapley values from game theory \cite{lundberg2017unified}. SHAP has gained considerable attention due to its solid theoretical foundation and its effectiveness in delivering reliable explanations across various domains, making it a widely accepted method for local explanations \cite{Fryer_2021}.



Despite the great success of SHAP, its application to large datasets is often hindered by its computational complexity. Calculating exact SHAP values is not only time-consuming but can also be impractical for many widely used models. To address these challenges, approximation methods such as Kernel SHAP have been proposed. Kernel SHAP estimates SHAP values by solving a weighted linear regression on a sample of perturbed instances, providing a more computationally efficient means of obtaining SHAP value estimates with fewer evaluations of the original model \cite{kelodjou2024shaping}. 

However, while Kernel SHAP offers improvements in efficiency, it introduces a new challenge: instability in explanations \cite{aas2021explaining}. Kernel SHAP, as well as its variants (Partitioning, and Sampling), depend on the choice of a background dataset as the baseline, which serves as a reference point for SHAP explanations and can strongly impact feature attribution values. Different executions of Kernel SHAP with the same inputs can produce varying explanations, leading to inconsistencies that can undermine the reliability of the results and decrease user trust \cite{aas2021explaining, chen2020algorithms}, particularly in anomaly detection for energy consumption data, where stable and consistent explanations are essential for effective decision-making and energy management.

To address these challenges, this paper proposes an approach for explaining anomaly detection models in energy consumption data that mitigates instabilities and enhances explanation reliability, particularly in SHAP-based techniques. The consistency of explanations is improved by selecting data samples relevant to the anomaly under observation when determining the baseline for SHAP value calculations. Moreover, our approach prioritizes the features most relevant to anomalies, providing consistent and accurate insights while minimizing computational burden by selecting only a few background data samples. Evaluation results across five datasets and 10 XAI techniques demonstrate improvements in explanation stability and reliability, with reduced variability in feature importance.

The rest of this paper is structured as follows: Section \ref{sec:RelatedWork} reviews related work, Section \ref{sec:Background} provides the background, Section \ref{sec:methodology} details the methodology, Section \ref{sec:Results} presents the results and discusses the finding, and Section \ref{sec:Conclusion} concludes the study.

\section{Related Work}
\label{sec:RelatedWork}
This section reviews XAI techniques in energy systems and time series applications, as well as in anomaly and fault detection.

\subsection{XAI in Energy Systems and Time Series Applications}
Several studies have leveraged XAI to provide insights into workings of energy load forecasting models and other building energy management systems. Moon et al. \cite{Moon_Rho_Baik_2022} proposed Explainable Electrical Load Forecasting (XELF) methodology for educational buildings, emphasizing the importance of understanding and interpreting the factors that influence electrical load predictions. By incorporating external factors such as weather and internal building data, they trained various tree-based models and utilized SHAP to provide interpretable explanations for the energy predictions made by these models. Similarly, Chung and Liu \cite{Chung_Liu_Analysisof_2022} analyzed input variables for deep learning models predicting building energy loads, comparing the XAI techniques Local Interpretable Model-Agnostic Explanations (LIME) and SHAP, and found that SHAP outperformed LIME by maintaining prediction accuracy with fewer input variables. Joshi et al. \cite{Joshi_Jana_Arjunan_2022} presented a data-driven approach for benchmarking energy usage in Singapore, employing ensemble tree models and XAI techniques such as SHAP for a detailed analysis of the impact of building attributes on energy consumption.

The integration of XAI in power systems and renewable energy has been also been explored. Zhang et al. \cite{Zhang_Xu_Zhang_2020} utilized SHAP to explain deep reinforcement learning models for power system emergency control, generating SHAP values to quantify the impact of each system variable and clarify how different factors influenced emergency control decisions. Tan et al. \cite{Tan_Zhao_Su_2022} proposed an explainable Bayesian neural network for probabilistic transient stability analysis in power systems, using the Gradient SHAP algorithm for explanations. Their approach provided insights at both global and local levels, with global explanations offering a comprehensive understanding of factors influencing overall model behavior and local explanations detailing individual predictions.

Leuthe et al. \cite{leuthe2024leveraging} explored XAI in building energy consumption forecasting and compared transparent and black-box models. They considered linear regression, decision tree, and QLattice as transparent prediction models and applied four XAI methods - partial dependency plots, Accumulated Local Effects (ALE), LIME, and SHAP - to an artificial neural network using a real-world dataset of residential buildings. Their findings indicate that appropriate XAI methods can significantly improve decision-makers' satisfaction and trust in machine learning models for energy forecasting. Mueller et al. \cite{mueller2023illuminating} examined the use of XAI to explain characteristics of vehicle power consumption, referring to the energy consumption within a vehicle's low-voltage electrical system. The study applied methods such as ALE and Permutation Feature Importance (PFI) for global insights, and LIME and SHAP for local analysis.

XAI techniques have also been investigated in time series analysis beyond the energy domain. Rožanec et al. \cite{Rožanec_Trajkova_Kenda_2021} examined the explainability of global time series forecasting models. They used the M4 competition and Kaggle Wikipedia Web Traffic datasets, integrating anomaly detection with XAI techniques to identify and explain poor forecasts. Anomaly detection flagged deviations, while XAI methods, such as LIME, computed feature attributions and generated counterfactual examples to elucidate the reasons behind these deviations. In addition, Labaien Soto et al. \cite{Labaien_Zugasti_De_2023} proposed a model-agnostic approach that uses autoencoders to generate real-time counterfactual explanations. This approach used one-dimensional Convolutional Neural Networks (1D-CNN) and Recurrent Neural Networks (RNNs) to analyze time series vibration data. The approach explains anomalies by making small input modifications to measure how far they deviate from normal behavior, providing insights into the factors contributing to the anomaly.

Freeborough and van Zyl \cite{Freeborough_van_InvestigatingExplainability_2022} explored XAI methods for financial time series forecasting, focusing on ablation, integrated gradients, added noise, and permutation techniques to assess feature importance and enhance model explainability. These methods were applied with several ML models including standard RNN, LSTM, and Gated Recurrent Unit (GRU). Schlegel and Keim \cite{Schlegel_Keim_ADeep_2023} focused on enhancing the explainability of deep learning models for time series data using perturbation analysis and evaluated state-of-the-art XAI techniques, including gradient-based methods (Saliency, Integrated Gradients) and SHAP-based methods (DeepLiftShap, Kernel SHAP), on three time-series classification datasets: FordA, FordB, and ElectricDevices. 
The study analyzed attribution distributions, skewness, and both Euclidean and cosine distances between original and perturbed instances. It found that SHAP and its derivatives generally produce effective attributions, while noting that the efficacy of XAI techniques can vary significantly depending on the chosen perturbation strategy.

Despite significant advancements in using XAI techniques for energy load forecasting and time series analysis, current methods do not adequately address the reliability of explanation results. Given that SHAP methods are grounded in tabular data, they needed to be adapted to accommodate time series data. SHAP struggles with computation complexity while SHAP variants result in unstable explanations. To address this, we propose a technique leveraging SHAP approaches, but address instability through targeted selection of the background dataset based on context-relevant information.

\subsection{XAI in Anomaly/Fault Detection}
XAI has been instrumental for enhancing the explainability of anomaly detection and fault detection models. Roshan and Zafar \cite{Roshan2021109} explored the use of SHAP for feature selection in an unsupervised anomaly detection setting. Their approach leveraged SHAP to improve the performance of autoencoders by identifying key features responsible for anomalies, and retraining the model using only benign data. While the SHAP-based feature selection showed improved results over other methods, the paper faced limitations in computational cost due to Kernel SHAP's complexity and potential sampling bias from using a subset of the CICIDS2017 dataset, which could affect generalizability. 
On the other hand, Antwarg et al. \cite{Antwarg2021} focused on applying the SHAP framework, traditionally used in supervised learning, to explain anomalies detected by unsupervised autoencoders. Their approach emphasizes understanding the relationships between features with high reconstruction errors and those most critical to anomaly detection. Kernel SHAP is utilized to calculate feature importance, offering detailed insights into why certain anomalies occur, by identifying both contributing and offsetting features. However, challenges such as selecting the appropriate background dataset and further validation across various autoencoder architectures remain areas for future research. While Roshan and Zafar's work primarily aimed at enhancing model prediction accuracy through feature selection, Antwarg et al. focused more on interpreting feature contributions and relationships in the context of anomaly explanations.

Kim et al. \cite{kim2021explainable} proposed an explainable anomaly detection framework for maritime main engine sensor data, combining SHAP with hierarchical clustering to interpret and group common anomaly patterns. By transforming SHAP values based on their distributions, Kim et al. were able to identify and isolate key sensor variables contributing to anomalies, allowing for more precise segmentation and analysis of the detected anomalies. This method provides insights into the causes of anomalies by visualizing and grouping similar patterns, offering an improved understanding of the conditions leading to engine failures. 
A two-layer patient monitoring system employing Kernel SHAP for anomaly detection and explanation in healthcare data was presented by Abououf et al. \cite{abououf2023explainable}. The most influential features contributing to anomalies detected by an autoencoder-based model were identified utilizing SHAP.

Kernel SHAP was also used for fault detection in industrial applications.
Asutkar and Tallur \cite{Asutkar2023} proposed an explainable unsupervised learning framework using autoencoders for fault detection and Kernel SHAP for explanations. Their framework accurately identified machine faults under varying operating conditions, with Kernel SHAP highlighting the most prominent features contributing to fault detection. The approach was validated using multiple datasets, demonstrating its effectiveness and scalability in real-world industrial applications. For process monitoring, Choi and Lee \cite{Choi2022182} proposed an explainable fault diagnosis model that combines Stacked Autoencoders (SAE) with Kernel SHAP for feature importance and model behavior explanation. Kernel SHAP provided a clear explanation of which features contributed most significantly to the model’s predictions. 

The Table \ref{tab:related_work_summary} summarizes key studies that have applied XAI techniques, especially SHAP variants, to energy consumption and related fields. These works highlight the benefits of using XAI methods to improve model transparency, interpretability, and decision-maker trust. For example, SHAP methods have been effectively used to analyze energy load forecasting, identify influential features, and provide both global and local explanations for model predictions. By incorporating XAI techniques, these studies have enhanced the understanding of how external factors such as weather, building attributes, and sensor data influence predictions, enabling more informed decision-making in energy systems.

\renewcommand{\arraystretch}{1.2} 

\afterpage{
{\fontsize{9}{10}\selectfont
\setlength{\tabcolsep}{4pt} 
\renewcommand{\arraystretch}{1.05} 

\begin{longtable}{|>{\raggedright\arraybackslash}p{1.6cm}|>{\raggedright\arraybackslash}p{6cm}|>{\raggedright\arraybackslash}p{3.8cm}|>{\raggedright\arraybackslash}p{3.8cm}|}
\caption{
\\
Summary of Related Works} \label{tab:related_work_summary} \\

\hline
\textbf{Reference} & \textbf{Main Points} & \textbf{Pros} & \textbf{Cons} \\ \hline
Moon et al. (2022) & Proposed XELF methodology for educational buildings using SHAP to interpret energy load predictions. & Improved transparency and accuracy. & Focused on educational buildings only, limited datasets. \\ \hline
Chung and Liu (2022) & Compared LIME and SHAP for building load prediction: SHAP reduced variables while maintaining accuracy. & Reduced input features, maintained accuracy, improved interpretability. & Limited to LIME and SHAP comparison, excluding other methods. \\ \hline

Joshi et al. (2022) & Benchmarked energy in Singapore buildings using ensemble trees and SHAP to analyze attributes. & Real data, enhanced interpretability, and benchmarking insights. & Limited to Singapore hotel and retail buildings. \\ \hline
Zhang et al. (2020) & Used SHAP to explain deep reinforcement learning models for power system emergency control. & Quantified variable impact, improved decision transparency. & Focused on power systems, high complexity of variable interpretation. \\ \hline
Tan et al. (2022) & An explainable Bayesian NN for probabilistic transient stability analysis in power systems using Gradient SHAP. & Global and local insights. & Lacks comparison with alternative methods and validation of result reliability. \\ \hline

Leuthe et al. (2024) & Explored XAI in building energy forecasting, applying XAI methods to ANN models with residential datasets. & Enhanced trust with transparent models, compared XAI techniques. & Limited to a single dataset, lacks diversity in building types. \\ \hline

Mueller et al. (2023) & Applied XAI methods (ALE, PFI, LIME, SHAP) to vehicle power consumption in low-voltage systems. & Combined global/local insights, enhanced transparency for complex systems. & Lacks comparison of reliability between global and local insights. \\ \hline


Rožanec et al. (2021) & Combined anomaly detection and XAI for time series forecasting, provided counterfactual explanations for poor forecasts. & Identified deviations, clarified anomalies with counterfactuals. & Lacks evaluation of broader explainability approaches. \\ \hline
Labaien Soto et al. (2023) & A model-agnostic approach with autoencoders for counterfactual explanations in anomaly detection. & Real-time explanations, works across model types. & Lacks comparison to diverse deep learning architectures. \\ \hline

Freeborough and van Zyl (2022) & XAI methods (e.g., ablation, integrated gradients, permutation) for financial time series with RNNs. & Effective feature attribution for RNNs, LSTMs, and GRUs. & Lacks assessment of XAI reliability and comparison across scenarios. \\ \hline


Schlegel and Keim (2023) & Deep learning explainability for time series using gradient- and SHAP-based methods across multiple datasets. & Provided detailed attribution insights, effective perturbation analysis. & Limited robustness evaluation and consistency analysis of XAI methods. \\ \hline

Roshan and Zafar (2021) & SHAP for feature selection in unsupervised anomaly detection with autoencoders. & Improved anomaly detection accuracy, highlighted key features. & High computational, sampling bias in datasets. \\ \hline


Antwarg et al. (2021) & SHAP to explain anomalies detected by unsupervised autoencoders, analyzing relationships between features. & Detailed insights into feature contributions and relationships. & Explanation stability and generalization across scenarios are unexamined. \\ \hline


Kim et al. (2021) & Developed an explainable anomaly detection framework for maritime engine data combining SHAP and clustering. & Identified key variables, segmented anomaly patterns. & Does not evaluate the robustness of explanations across anomaly patterns.\\ \hline
Abououf et al. (2023) & Proposed a two-layer patient monitoring system using Kernel SHAP for anomaly detection in healthcare data. & Improved anomaly interpretability for healthcare monitoring. & High computational cost, limited to specific healthcare applications. \\ \hline

Asutkar and Tallur (2023) & Used Kernel SHAP in an unsupervised framework for fault detection in industrial machines. & Effective under varying conditions, scalable for Industry 4.0. & Limited exploration of explainability in multi-fault scenarios. \\ \hline
Choi and Lee (2022) & Combined SAE and Kernel SHAP for explainable fault diagnosis in process monitoring. & Clear feature importance, high classification accuracy. & Limited to specific industrial process data. \\ \hline
\end{longtable}
}
}

Although XAI methods such as SHAP have been successfully applied to anomaly detection and fault diagnosis, particularly with autoencoders, a gap remains in ensuring the reliability and stability of explanations. Current approaches suffer from inconsistencies, as noted in multiple studies \cite{Roshan2021109, Antwarg2021}, due to the random sampling of training data for the background dataset. For instance, Chung and Liu \cite{Chung_Liu_Analysisof_2022} noted variability in SHAP explanations due to random background data selection and input dependencies, while Antwarg et al. \cite{Antwarg2021} and Chen et al. \cite{chen2020algorithms} highlighted challenges in selecting appropriate background datasets, which significantly influences explanation stability. Similarly, Schlegel and Keim \cite{Schlegel_Keim_ADeep_2023} reported inconsistencies tied to perturbation strategies, often exacerbated by arbitrary baseline datasets, and Roshan and Zafar \cite{Roshan2021109} documented issues with sampling bias impacting explanation robustness. We address these issues by proposing a robust approach that provides stable explanations focusing on highly relevant data for the anomaly under consideration. By aligning baselines with the contextual data of anomalies, our method provides consistent and reliable insights, representing a significant improvement over previous practices and enhancing the practical utility of explainability in anomaly detection.

\section{Background}
\label{sec:Background}
This section first introduces core concepts in explaining black box models. Next, classical SHAP and Kernel SHAP are introduced, as our approach leverages these techniques.

\subsection{Explaining Black-Box Models}

Understanding and interpreting the decisions made by Artificial Intelligence (AI) models, especially those that function as black-box systems, is crucial for ensuring their legitimacy and reliability in sensitive applications \cite{goodman2017european}. Modern AI models such as Deep Neural Networks (DNNs) are complex systems with many parameters, making them difficult to interpret \cite{castelvecchi2016can}. To address this challenge XAI has emerged, offering two key types of explainability: local and global. Local explainability focuses on explaining the decision for a specific instance, offering detailed insights into how a particular prediction was made. In contrast, global explainability provides an overarching view of the model's decision-making process, giving a broader understanding of the AI system's behavior across its entire input space \cite{spyrison2024exploring, fernandez2024flocalx}. Both local and global explainability play important roles in making AI systems more transparent, accountable, and understandable, thereby enhancing trust and facilitating decision-making \cite{fernandez2024flocalx}.


In this research, we focus on local explainability because it allows for a deeper understanding of individual anomalies detected within the data, assisting in determining the root cause of each anomaly. Both LIME and SHAP provide local explanations. LIME explains one prediction at a time by constructing a simple linear model around the data point, using random perturbation to create simulated data. However, this can lead to instability of explanations. Conversely, SHAP provides individual explanations by assigning feature importance based on Shapley values, offering a more stable approach \cite{zafar2021deterministic}.

\subsection{Classic SHAP and Kernel SHAP}

Shapley value estimation, based on cooperative game theory, calculates the contribution of each feature by comparing the model's predictions with and without that feature across all possible feature combinations \cite{lundberg2017unified}.
However, this is computationally expensive, especially with large feature sets. To address this, sampling methods approximate Shapley values without requiring retraining for every combination of features \cite{lipovetsky2001analysis}. Despite these approximations, the process remains resource-intensive for large datasets, leading to the development of more efficient methods such as Kernel SHAP.



Building on the classic Shapley value estimation, Kernel SHAP provides a more practical approach to approximate Shapley values by solving a Weighted Least Squares (WLS) problem. This weighted approach assigns different importance levels to each possible subset of features. Kernel SHAP utilizes sampling techniques to approximate Shapley values while reducing computational burden, thus making it feasible for models with many features while maintaining their theoretical integrity \cite{lundberg2017unified, aas2021explaining}. This balance of practicality and rigor makes Kernel SHAP well suited for energy applications; nevertheless, it still suffers from instability.
The Kernel SHAP  estimates the contribution of each feature through the following WLS problem:

\begin{equation} \label{eq:kernel_shap}
\min_{\phi_0, \ldots, \phi_{F}} \sum_{S \subseteq F} k(F, S) \left( v(S) - \left( \phi_0 + \sum_{j \in S} \phi_j \right) \right)^2
\end{equation}


Here \( F \) represents the total number of features in the model, and \( S \) denotes a subset of features. The term \( v(S) \) refers to the value of the predictive model when only the features in subset \( S \) are considered. The term \( \phi_0 \) is the base value representing the average prediction over the entire background dataset. The contribution of feature \( j \) indicated by \( \phi_j \) is calculated as part of the WLS optimization process by balancing the error terms across all feature subsets.
The weights \( k(F, S) \) in Equation \ref{eq:kernel_shap}, are given by:

\begin{equation} \label{eq:kernel_weights}
k(F, S) = \frac{(|F| - 1)}{\binom{|F|}{|S|} \cdot |S| \cdot (|F| - |S|)}
\end{equation}


Here \( \binom{|F|}{|S|} \) is the binomial coefficient representing the number of ways to choose \( |S| \) features from the total \( |F| \) features. A subset \( S \) refers to any possible combination of features from the full set \( F \). For example, given three features \( \{A, B, C\} \), possible subsets include \( \{A\}, \{B, C\}, \{A, B, C\}, \) and the empty set \( \{\} \). The kernel weights \( k(F, S) \), defined in Equation \ref{eq:kernel_weights}, are derived from cooperative game theory, where the contribution of each feature is evaluated by considering all possible subsets \( S \) of features. These weights \( k(F, S) \) ensure that each feature’s contribution is assessed in a balanced manner by giving different levels of importance to subsets of different sizes. More specifically, the weighting ensures that features are not biased by their position in a subset and that smaller subsets are given fair consideration \cite{aas2021explaining}. This weighting scheme balances the evaluation, ensuring that each feature’s contribution is assessed within a meaningful and fair context.

Both Classic SHAP and Kernel SHAP determine feature importance by excluding features from the model to observe their impact on predictions. Since the model is already trained and features cannot be physically removed, features are substituted with alternative values to reduce their influence. This substitution typically involves using values from a designated background dataset. When calculating the model's prediction without a specific feature, the real value of that feature is replaced with a value from the background dataset. The background dataset can consist of either the entire training data or, in the case of large datasets, a representative subset. This allows the model to simulate the absence of the feature and assess its contribution by comparing changes in predictions when the feature is excluded.

\section{Methodology}
\label{sec:methodology}

\begin{figure*}[b!]
    \centering
    \includegraphics[width=1\textwidth]{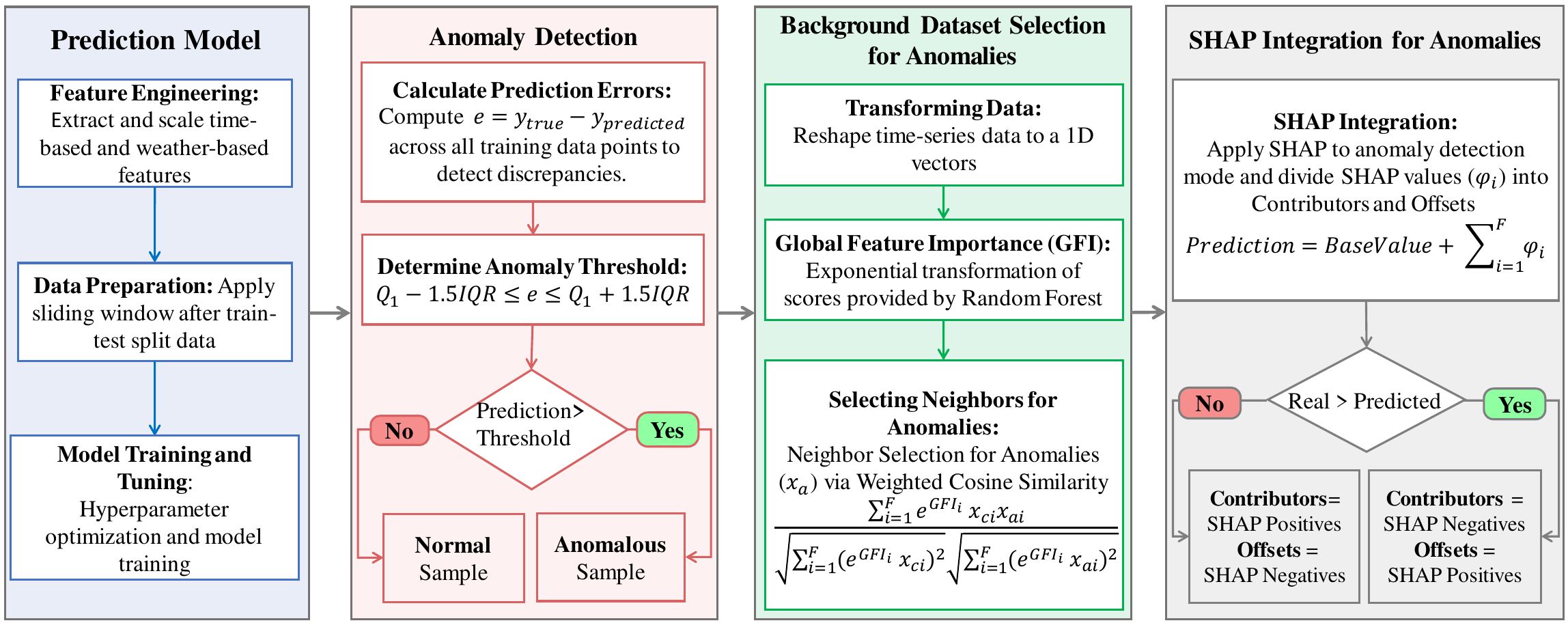}
    \caption{Methodology overview}
    \label{fig:methodology-overview}
\end{figure*}

This section presents the proposed approach for explaining the anomaly detection model for energy consumption data by leveraging variants of SHAP, such as Kernel SHAP, but improving stability and consistency through the targeted selection of the background dataset (baseline) using a weighted cosine similarity technique. The approach is specifically designed for prediction-based anomaly detection techniques where a black-box model is used to generate energy predictions, which are in turn compared to actual energy values. If the difference exceeds the threshold, the sample is deemed anomalous. After detecting anomalies, we proceed to the explanation phase, which involves multiple steps to provide clear feature contributions for each anomaly. The overview of the complete process of detecting anomalies and explaining underlying features is presented in Figure \ref{fig:methodology-overview}, while the details of each component are presented in the following subsections.

\subsection{Prediction Model}


This first step involves getting the prediction-based model ready for the anomaly detection. As seen from Figure \ref{fig:methodology-overview}, it consists of feature engineering, data preparation, and model training and tuning.

\subsubsection{Feature Engineering} \label{sec:FeatureEngineering}


Energy consumption data obtained from smart meters or other sensors typically consists of energy consumption and the reading date-time recorded in hourly or similar intervals. For better anomaly detection, we extract the following features from the reading date-time: the hour of the day, the day of the week, the day of the month, the day of the year, the month, and an indicator for weekends. Weather-related features such as temperature, humidity, and wind speed are also incorporated, along with previous energy consumption readings as input features, while the energy consumption remains the target variable. Data are scaled using Min-Max scaling to bring all features to a similar range and avoid dominance of large numbers:   

\begin{equation}
X' = \frac{X - X_{\mathrm{min}}}{X_{\mathrm{max}} - X_{\mathrm{min}}}
\label{eq:min_max_scaler}
\end{equation}

\noindent where \( X \) represents the original feature values, \( X_{\mathrm{min}} \) and \( X_{\mathrm{max}} \) are the minimum and maximum values of the feature, and \( X' \) are the scaled values.

\subsubsection{Data Preparation}

\begin{figure}[b!]
    \centering
    \includegraphics[width=0.8\linewidth]{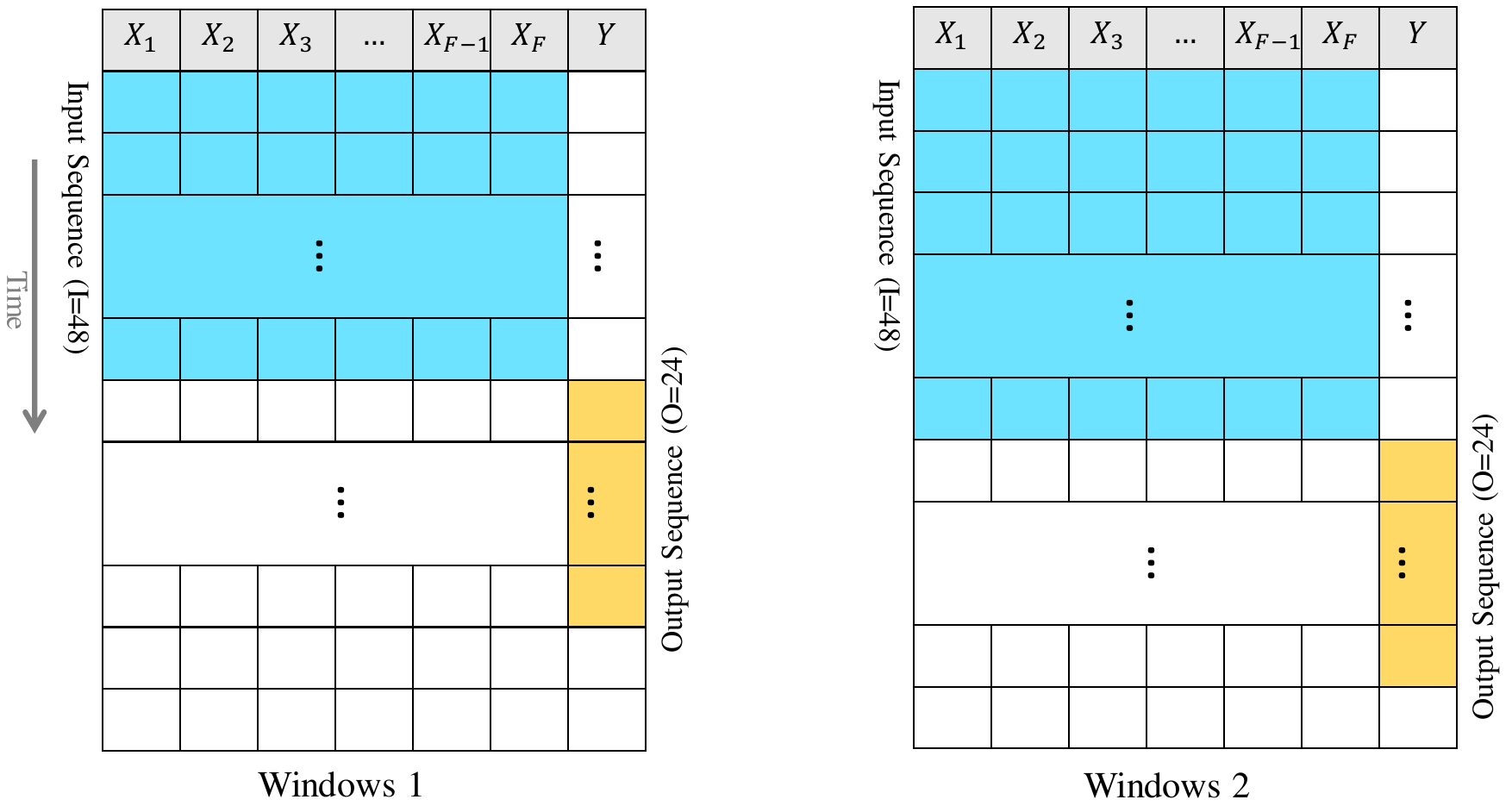}
    \caption{Illustration of the sliding window approach for sequence-to-sequence model training with 48 Input sequence and 24 output sequence}
    \label{fig:seqbyseq_data_format}
\end{figure}

Next, the dataset is chronologically split, with the first 80\% used for training, the next 10\% for validation, and the final 10\% for testing. To capture temporal dependencies, the sliding window technique is applied, moving a fixed-length window along the time series, advancing one record at a time. As shown in Figure \ref{fig:seqbyseq_data_format}, this approach creates sequences of inputs and corresponding outputs for the model to learn temporal relationships. For each window \( i \), the input sequence is represented as a matrix of time steps and features:

\begin{equation}
\begin{bmatrix}
X_{1, t_1} & X_{2, t_1} & \dots & X_{F, t_1} \\
X_{1, t_2} & X_{2, t_2} & \dots & X_{F, t_2} \\
\vdots & \vdots & \ddots & \vdots \\
X_{1, t_I} & X_{2, t_I} & \dots & X_{F, t_I} \\
\end{bmatrix}
\label{eq:input_sequence}
\end{equation}


\noindent where \( I \) represents the number of time steps in the sliding window, \( F \) is the number of features, and \( X_{f,t} \) is the value of feature \( f \) at time \( t \).

The output sequence for window \( i \) is represented as:

\begin{equation}
O_i = [Y_{t+1}, Y_{t+2}, \dots, Y_{t+h}]
\label{eq:output_sequence}
\end{equation}


\noindent where \( h \) is the forecasting horizon and \(  [Y_{t+1}, Y_{t+2}, \dots, Y_{t+h}] \) are the target energy consumption values for time steps \( t+1 \) to \( t+h \). In this paper, we used four time steps  (\( I=48 \)) for inputs and one step ahead forecasting horizon (\( h=24 \)) as shown in Figure \ref{fig:seqbyseq_data_format}, but explanation only provided for first horizon (\( h=1 \)). The prediction model supports anomaly detection through comparison of predicted and actual values.

\subsubsection{Model Training and Tuning}


Here, the prediction model, regardless of which type of architecture is employed, is trained and tuned. The selection of hyperparameters that need to be tuned depends on the selected model, but the tuning process remains the same. Bayesian optimization, specifically using the Tree-structured Parzen Estimator (TPE) \cite{akiba2019optuna}, is employed for tuning due to its resource efficiency, although other techniques could be used as well. Tuning is carried out with the validation set.


The optimization process starts with defining the hyperparameter search space, outlining the ranges and types of hyperparameters to be optimized. The TPE algorithm is then initialized, and the model performance is assessed based on the initial hyperparameters. After each evaluation, the model updates the hyperparameter probabilities, and new, more promising hyperparameters are selected. This cycle of evaluation and refinement continues until the optimal set of hyperparameters is found.
Once hyperparameters are selected through TPE, the model is trained using the these hyperparameters on the training data. Finally, the trained prediction model is ready for anomaly detection.

\subsection{Anomaly Detection}


As already mentioned, prediction-based anomaly detection methods, including those employed in this study, identify the anomalies by comparing the predicted values with the actual values. Normal data is expected to have small deviations, while anomalous samples are expected to result in larger discrepancies. As seen from Figure \ref{fig:methodology-overview}, anomaly detection involves calculating prediction error, determining an anomaly threshold, and classifying samples as  anomalous or non-anomalous. 




Next, we calculate the \textit{prediction error} for the entire training dataset. The prediction error is defined as the difference between the actual and predicted values for each data point within a single prediction window. 
By computing the prediction errors across all data points in the training dataset, we obtain a comprehensive distribution of errors. 



\begin{figure*}[b!]
    \centering
    \includegraphics[width=0.7\linewidth]{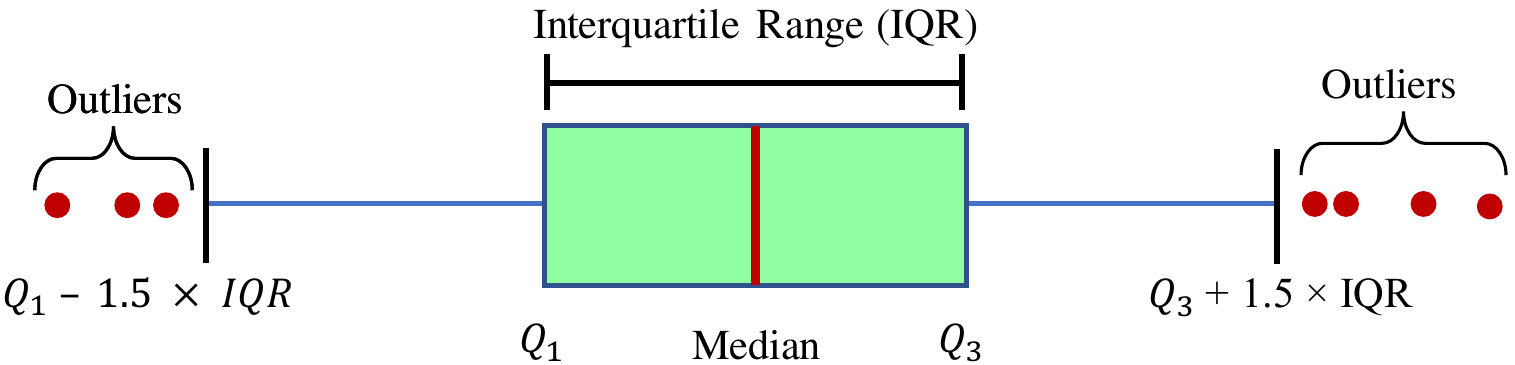}
    \caption{Illustration of the anomaly detection process using IQR}
    \label{fig:IQR}
\end{figure*}

Next, the anomaly threshold is determined using the Interquartile Range (IQR) method calculated based on the prediction errors from all training data, as it is well-suited for identifying outliers in skewed data distributions \cite{dallah2022outlier}. The IQR method sets a range that defines normal data behavior, as shown in Equation \ref{eq:iqr_method}:

\begin{equation}
Q_1 - 1.5 \times IQR \leq e \leq Q_3 + 1.5 \times IQR
\label{eq:iqr_method}
\end{equation}


\noindent where \(Q_1\) represents the first quartile (25th percentile) of the prediction errors, \(Q_3\) is the third quartile (75th percentile), and \(IQR\) is the interquartile range, calculated as \(Q_3 - Q_1\).
Figure \ref{fig:IQR} illustrates this anomaly detection process using the IQR method. 

Each prediction error is compared against the established threshold. If the prediction error exceeds this threshold, the data point is classified as an \textit{Anomalous Sample}, suggesting that it deviates significantly from expected behavior. Conversely, if the prediction error falls within the threshold, the data point is classified as a \textit{Probably Normal Sample}. This classification enables the identification of potential issues in energy consumption.

\subsection{Background Dataset Selection for Anomalies} \label{sec:main}

In the context of SHAP value explanations, the background dataset serves as the baseline, representing the expected or average behavior against which individual predictions are compared. According to Chen et al. \cite{chen2020algorithms}, selecting an appropriate baseline is crucial as it defines how absent feature values are handled during Shapley value calculations. Different baseline strategies, such as fixed baselines or distributional baselines (marginal and conditional), influence the resulting SHAP values in distinct ways. Our approach of selecting a similar background dataset using weighted cosine similarity aligns with the concept of distributional baselines, where the background samples are chosen to reflect the statistical context of the anomalies being explained. By carefully selecting neighbors that closely resemble the anomaly points, we ensure that the baseline accurately captures the relevant feature distributions and dependencies, thereby enhancing the interpretability and reliability of the SHAP value explanations. This method mitigates the arbitrariness associated with fixed baselines and preserves the integrity of feature relationships, particularly in complex, high-dimensional datasets. The process is presented in Algorithm \ref{alg:anomaly_explanation} and detailed in the following subsections.

\subsubsection{Transforming Data}


As our approach leverages the SHAP algorithm, it is necessary to first transform the windowed time series data into a format compatible with this method, which operates on tabular datasets with row vectors as samples. This transformation process is the first step of Algorithm~\ref{alg:anomaly_explanation}. For anomaly detection, the energy data is represented in a windowed format as shown in Equation \ref{eq:input_sequence}. These data are transformed into a tabular format \((1 \times I \times F)\), representing the multi-dimensional input into a one-dimensional vector. The transformed data are expressed as follows:


\begin{equation}
\begin{bmatrix}
X_{\mathrm{11}}, X_{12}, \dots, X_{\mathrm{1}F}, X_{\mathrm{21}}, X_{\mathrm{22}}, \dots, X_{\mathrm{2}F}, \dots, X_{IF}
\end{bmatrix}_{(1 \times I \times F)}
\label{eq:transformed_sequence}
\end{equation}


For example, when the model uses inputs with a window length of 48 and 10 features, this transformation results in a 480-dimensional vector representing the input, with explanations provided for each feature.


This paper focuses on explaining one-step-ahead predictions, corresponding to the first horizon of the output sequence \( O_i = [Y_{t+1}] \). By setting the output window \( h \) to 1, this approach facilitates early anomaly detection. However, the same approach can be extended to explain predictions for multiple horizons. To do this, one only needs to substitute the first horizon with the desired horizon (e.g., the second horizon) and repeat the same transformation and evaluation steps to provide explanations for subsequent horizons.

\subsubsection{Global Feature Importance} \label{sec:globalfeatureImportance}


After data are transformed into one-dimensional vectors, a surrogate model is used to calculate global feature importance (step 2 of  of Algorithm~\ref{alg:anomaly_explanation}). While various models can serve this purpose, we selected Random Forest due to its speed and computational effectiveness. Instead of using approaches that focus primarily on local explanations, such as SHAP, we employed Random Forest to provide faster global feature importance estimates. This is because it uses Gini importance, which does not require calculating local feature importance for each individual training data point. This advantage makes it well-suited for large datasets while maintaining high accuracy \cite{nembrini2018revival}. Moreover, Random Forest's ability to handle high-dimensional datasets and robust performance across various datasets made it the an excellent choice for this anomaly explanation step \cite{Wang_2016}.


To further strengthen the feature importance derived from the Random Forest model, the exponential transformation is applied to the importance scores, defined as \( GFI_i' = \exp(GFI_i) \). This transformation increases the differences in importance among features, giving more important features greater impact in subsequent analyses. As a result, the exponential scaling makes similarity measures, specifically cosine similarity, pay more attention to important features, improving the selection of relevant neighbors for SHAP explanations.

\subsubsection{Selecting Neighbors for Anomalies}


In step 3 of Algorithm~\ref{alg:anomaly_explanation}, weighted K-Nearest Neighbors (KNN) is employed to select neighbors for each detected anomaly point under analysis. The similarity score between an anomaly point \( x_a \) and each point \( x_c \) in the training set is computed using the weighted cosine similarity metric:

\begin{equation}
S(x_c, x_a) = \frac{\sum_{i=1}^{F} GFI_i' \cdot x_{ci} \cdot x_{ai}}{\sqrt{\sum_{i=1}^{F} (GFI_i' \cdot x_{ci})^2} \cdot \sqrt{\sum_{i=1}^{F} (GFI_i' \cdot x_{ai})^2}}
\label{eq:cosine_similarity}
\end{equation}


Here, \( x_{ai} \) and \( x_{ci} \) represent the values of feature \( i \) for the anomalous sample and another sample from the training dataset, respectively, and \( F \) denotes the total number of features. The wight of feature \( i \), \( GFI_i' \), is the global feature importance calculated determined as described in Subsection \ref{sec:globalfeatureImportance}. For each anomaly point, the top 100 most similar samples are initially selected as the background dataset for calculating local SHAP values. Future work may explore the impact of varying the number of selected samples on both computational efficiency and the quality of SHAP explanations to further optimize the process.

\begin{algorithm}
\caption{Background Dataset Selection for Anomaly Explanation}
\begin{algorithmic}[1]
\State \textbf{Input:} Training data $X_{\mathrm{train}}$, Test data $X_{\mathrm{test}}$, Trained model $M$
\State \textbf{Output:} Background dataset $B$ for anomaly explanation

\State \textbf{Step 1: Transform Windowed Data into 1D Format} \label{alg:step1}
\State Reshape windowed time series data from $X \in \mathbb{R}^{N \times T \times F}$ to $X' \in \mathbb{R}^{N \times (T \times F)}$

\State \textbf{Step 2: Calculate Global Feature Importance (GFI)} \label{alg:step2}
\State Train Random Forest regression on $X_{\mathrm{train}}$ to obtain feature importances $GFI_i$ for each feature $i$

\State \textbf{Step 3: Select Neighbors for Anomaly Points} \label{alg:step3}
\For{each anomaly point $x_a \in X_{\mathrm{test}}$}
    \For{each candidate point $x_c \in X_{\mathrm{train}}$}
        \State Compute weighted cosine similarity:
        \State \quad $S(x_c, x_a) = \dfrac{\sum_{i} GFI_i' x_{ci} x_{ai}}{\sqrt{\sum_{i} GFI_i' x_{ci}^2} \sqrt{\sum_{i} GFI_i' x_{ai}^2}}$
         \State where weights \( GFI_i' = \exp(GFI_i) \)
    \EndFor
    \State Select top $K$ candidate points with highest $S(x_c, x_a)$
    \State Set $B_{x_a} = \{ x_c \,|\, x_c \text{ is among top } K \text{ candidates} \}$
\EndFor


\end{algorithmic}
\label{alg:anomaly_explanation}
\end{algorithm}

\subsection{SHAP Integration for Anomalies}

After the SHAP values are calculated, features need to be categorized into those that contribute to the anomaly (referred to as \textbf{contributors}) and those act as offsets. This categorization is based on the comparison between the actual observed output and the predicted value for a specific instance. Such categorization based on SHAP values help us understand how much each feature pushes the model’s prediction away from or towards the real value. The prediction for a sample is expressed as \cite{aas2021explaining}:
 

\begin{equation} \label{eq:prediction}
\mathrm{Prediction} = \mathrm{Base Value} + \sum_{i=1}^{F} \phi_i
\end{equation}


Here, the term \textbf{BaseValue} represents the average prediction made using the background dataset, and the SHAP value \( \phi_i \) explains how much feature \( i \) contributes to the prediction. It is important to note that SHAP values are not used during the training of the model but are computed afterward to explain how features influence the final prediction.

The features are categorized according to the following logic:

\begin{itemize}
    \item \textbf{If Real \( > \) Predicted}: This indicates that the model under-predicted the outcome. In this case:
    \begin{equation}
    \mathrm{Real} > \mathrm{Base Value} + \sum_{i=1}^{F} \phi_i
    \label{eq:under_predict}
    \end{equation}

    Here, positive SHAP values (\( \phi_i > 0 \)) indicate features that pull the prediction closer to the real value, thereby acting as \textbf{offsets}. Negative SHAP values (\( \phi_i < 0 \)) indicate features that push the prediction further away from the true value, and these are considered \textbf{contributors} to the anomaly.

    \item \textbf{If Predicted \( > \) Real}: This indicates that the model over-predicted the outcome. In this case:
    \begin{equation}
    \mathrm{Base Value} + \sum_{i=1}^{F} \phi_i > \mathrm{Real}
    \label{eq:over_predict}
    \end{equation}
    Positive SHAP values (\( \phi_i > 0 \)) represent features that push the prediction further from the real value, contributing to the over-prediction (\textbf{contributors}). Conversely, negative SHAP values (\( \phi_i < 0 \)) pull the prediction closer to the real value, acting as \textbf{offsets}.
\end{itemize}



In the previous steps, we reduced the contribution of many less effective features to near-zero values. However, some features still require further filtering, as shown in Figure 7. 
By applying this categorization, we focus on features with negative SHAP values, which contribute to the anomaly. In this case, since the prediction is lower than the true value, the negative features (blue) are the main contributors, allowing us to disregard the positive features (red) for a more streamlined analysis.
This process of classifying SHAP values is a part of SHAP integration for anomalies process as illustrated in Figure \ref{fig:methodology-overview}.

\section{Results and Discussion} \label{sec:Results}

This section present the outcomes of our explainable anomaly detection approach, including data preparation, model optimization, prediction performance, and feature importance. The results highlight the improvements achieved through hyperparameter tuning, the impact of key features on model predictions, and the advantages of using our approach for enhancing model explainability. Statistical comparisons further support the robustness of our approach, demonstrating its effectiveness in producing reliable and consistent results.

\subsection{Dataset Description and Preparation}

To conduct a comprehensive analysis of the proposed method, we use energy consumption datasets from five different consumer types: a residence, a manufacturing facility, a medical clinic, a retail store, and an office building. A residence dataset provided by London Hydro \cite{londonhydro}, comprises energy consumption records from a residence in London, Ontario, Canada, spanning from January 1, 2002, to December 31, 2004, with hourly energy consumption values. To enhance the predictive
capabilities of anomaly detection and improve explanations as discussed in Subsection \ref{sec:FeatureEngineering}, we incorporated additional
date-time and weather-related features from the Government of Canada’s historical climate data \cite{weather_canada_historical_data}. The remaining four datasets are from Building Data Genome Project 2 \cite{Miller2020yc}, covering January 2016 to December 2017.  Weather-related information for these datasets was already included in the repository.

The dataset was divided into training, validation, and test sets, following an 80-10-10 split. The model was trained based on windows sequences data. This configuration captures temporal dependencies and patterns within the data, significantly contributing to accurate anomaly detection in the analysis.


\subsection{Optimization and Training Results}

This study employs various deep learning architectures, including different variants of Recurrent Neural Networks (RNNs), Convolutional Neural Networks (CNNs), and Transformer-based models. The RNN variants utilized  are Long Short-Term Memory (LSTM), Gated Recurrent Unit (GRU), and their bidirectional counterparts (BLSTM and BGRU), which effectively capture temporal dependencies in sequential data. Within the CNN category, we used one-dimensional CNN (1D-CNN), Dilated CNN (DCNN), Temporal Convolutional Networks (TCN), and WaveNet, which efficiently capture local temporal patterns and enhance feature extraction through convolutional operations. Additionally, Transformer-based models such as the Temporal Fusion Transformer (TFT) and Time Series Transformer (TST) were employed, leveraging self-attention mechanisms to dynamically weigh the importance of time steps and features.




Hyperparameter optimization was conducted for all algorithms using the Tree-structured Parzen Estimator (TPE) from Bayesian optimization. A total of 50 trials were performed, with each trial representing a different set of hyperparameters evaluated on the validation set. The best combination was selected to minimize validation loss. A summary of the selected hyperparameters for models trained on the residential dataset and performance metrics is presented in Table \ref{tab:PerformanceMetrics}. For the remaining datasets, the same process was followed, but selected hyperparameters are not included for conciseness and  Table \ref{tab:accuracyOn4Datasets} only includes performance metrics.

\renewcommand{\arraystretch}{1.3}  
\begin{table*}[!b]
    \centering
    \setlength{\tabcolsep}{15pt}
    \caption{
    \\
    Summary of best hyperparameters and performance metrics for various models for the residential dataset.}
    \vspace{-5pt}
    \fontsize{9}{9.5}\selectfont  
    \resizebox{\textwidth}{!}{%
\begin{tabular}{l@{\hspace{0.15cm}} m{7cm} c@{\hspace{0.15cm}} c@{\hspace{0.15cm}} c@{\hspace{0.15cm}} c@{\hspace{0.15cm}} c@{\hspace{0.15cm}} c@{\hspace{0.15cm}}}
        \toprule
        \textbf{Model} & \textbf{Best Parameters} & \textbf{MSE} & \textbf{RMSE} & \textbf{MAE} & \textbf{SMAPE} & \textbf{MAPE} & \textbf{R\textsuperscript{2}} \\
        \midrule
        LSTM & 88 LSTM units, dropout rate 0.224, 2 LSTM layers, learning rate 0.000279 & 0.09 & 0.30 & 0.15 & 9.23 & 8.98 & 0.41 \\
        GRU & 88 GRU units, dropout rate 0.257, 1 GRU layer, learning rate 0.000997 & \textbf{0.07} & \textbf{0.26} & \textbf{0.14} & \textbf{9.04} & \textbf{8.84} & \textbf{0.53} \\
        BLSTM & 109 LSTM units, dropout rate 0.285, 2 LSTM layers, learning rate 0.000189 & 0.09 & 0.30 & 0.15 & 9.18 & 8.96 & 0.43 \\
        BGRU & 76 GRU units, dropout rate 0.272, 1 GRU layer, learning rate 0.000907 & 0.10 & 0.32 & 0.22 & 14.37 & 15.59 & 0.33 \\
        CNN & 57 filters, kernel size 4, dropout rate 0.348, Adam optimizer, L2 regularization 1.01e-06 & 0.08 & 0.28 & 0.16 & 10.20 & 10.35 & 0.45 \\
        TCN & 69 filters, kernel size 5, dropout rate 0.400, 5 TCN blocks, dilation base 3, L2 regularization 1.23e-06, Adam optimizer & 0.08 & 0.28 & 0.17 & 10.95 & 11.41 & 0.46 \\
        DCNN & 42 filters, kernel size 3, dilation rate 4, dropout rate 0.115, 1 convolutional layer, RMSprop optimizer, L2 regularization 7.37e-05 & 0.08 & 0.28 & 0.16 & 10.33 & 10.61 & 0.48 \\
        WaveNet & 61 filters, kernel size 2, dilation rate 4, 2 WaveNet blocks, dropout rate 0.386, L2 regularization 7.84e-04, Adam optimizer & \textbf{0.07} & \textbf{0.26} & \textbf{0.14} & \textbf{9.04} & 9.12 & \textbf{0.53} \\
        TFT & 255 hidden units, dropout rate 0.312, 1 LSTM layer, 3 attention heads, 3 attention blocks, L2 regularization 8.81e-06, learning rate 0.000144, Adam optimizer & 0.12 & 0.35 & 0.19 & 12.13 & 12.18 & 0.22 \\
        TST & Model dimension 65, 4 attention heads, 4 transformer layers, feed-forward dimension 471, dropout rate 0.306, learning rate 5.54e-05, Adam optimizer & 0.11 & 0.33 & 0.22 & 13.79 & 14.69 & 0.29 \\
        \bottomrule
    \end{tabular}
    }
    \label{tab:PerformanceMetrics}
\end{table*}
\renewcommand{\arraystretch}{1}  

\begin{figure*}[t]
    \centering
    \scalebox{0.4}{\includegraphics{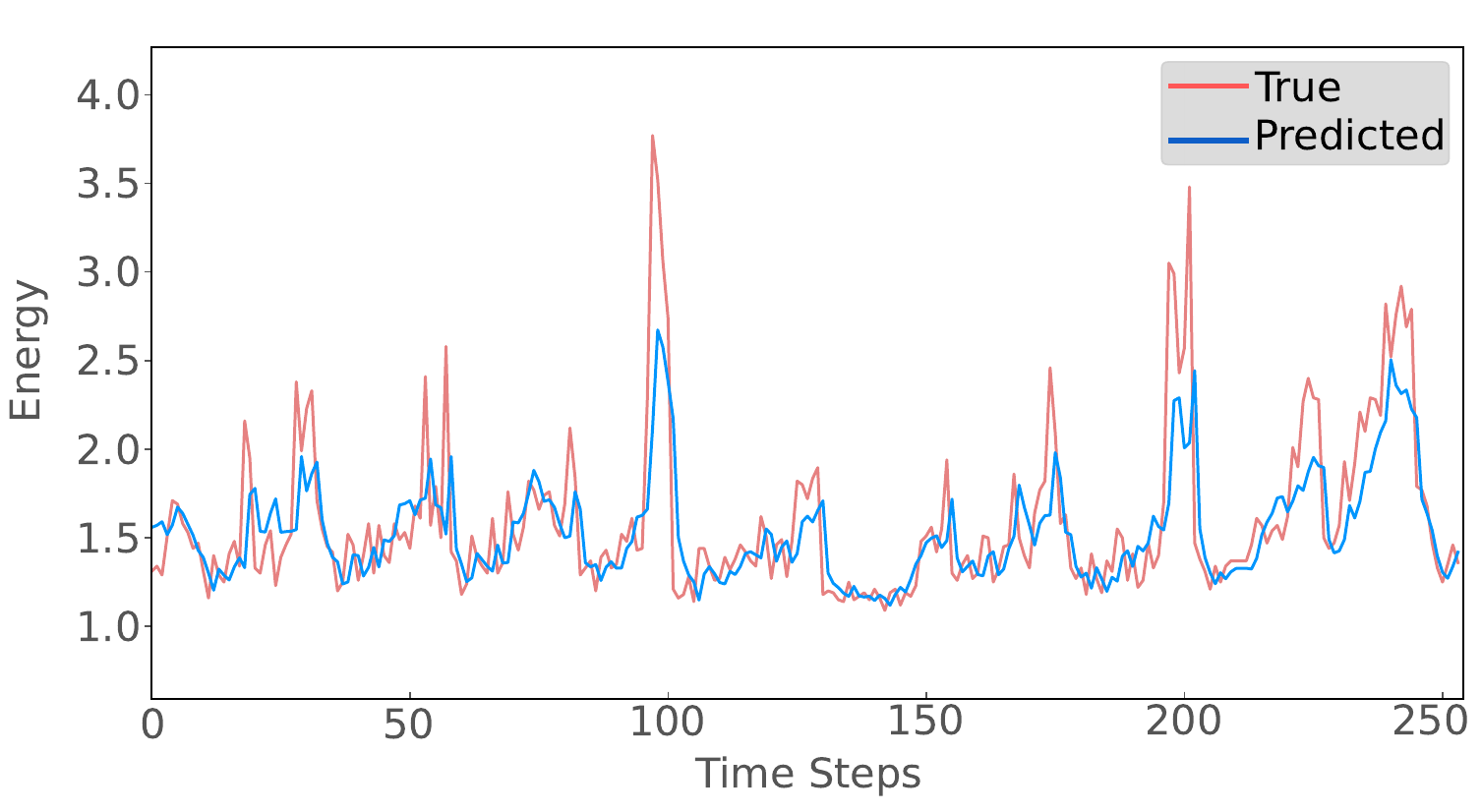}}
    \caption{Prediction results for the LSTM model, showing 10\% of the test residential dataset.}
    \label{fig:LSTM_PerformancePlot}
\end{figure*}

Next, the models with the best hyperparameter combinations were selected, and their performance was evaluated on the test set. Evaluation metrics included Mean Squared Error (MSE), Root Mean Squared Error (RMSE), Mean Absolute Error (MAE), Symmetric Mean Absolute Percentage Error (SMAPE), Mean Absolute Percentage Error (MAPE), and \( R^{2} \). 
Table \ref{tab:PerformanceMetrics} highlights the reliability of the prediction models on the residential dataset, which is essential for generating trustworthy explanations. Inaccurate models lead to unreliable explanations. The table presents various deep learning models with optimized hyperparameters, ensuring a fair performance comparison. Most models achieved high accuracy with low error rates, particularly the GRU model, which demonstrated the lowest errors and highest R\textsuperscript{2} score. 
Figure \ref{fig:LSTM_PerformancePlot} shows an example of the LSTM model predictions compared to the actual values on 10\% of the test data from the residential dataset, showing alignment between predicted and actual energy consumption.

\renewcommand{\arraystretch}{1.2}  
\noindent \begin{table*}[!b]
\centering
\caption{\\
Summary of performance metrics for additional datasets.}
\setlength{\tabcolsep}{1.5pt}  
\renewcommand{\baselinestretch}{1}  
\fontsize{8.5}{9}\selectfont  
\noindent \begin{tabular}{cc}
\begin{minipage}{0.48\linewidth}
\fontsize{8.5}{9}\selectfont  
\centering
\caption*{\textbf{(a) Manufacturing Facility}}
\vspace{-5pt}
\begin{tabular}{l c c c c c c}
\hline
\textbf{Model} & \textbf{MSE} & \textbf{RMSE} & \textbf{MAE} & \textbf{SMAPE} & \textbf{MAPE} & \textbf{R\textsuperscript{2}} \\
\hline
LSTM & 254.49 & 15.95 & 9.18 & 11.21 & 11.42 & 0.9 \\
GRU & 306.9 & 17.52 & 11.03 & 14.33 & 14.5 & 0.88 \\
BLSTM & 291.82 & 17.08 & 9.93 & 12.13 & 12.63 & 0.88 \\
BGRU & 318.79 & 17.85 & 10.6 & 12.96 & 13.08 & 0.87 \\
CNN & 201.35 & 14.19 & 7.85 & 9.89 & 10.15 & \textbf{0.92} \\
TCN & 200.1 & 14.15 & 8.16 & 10.31 & 10.9 & \textbf{0.92} \\
DCNN & 209.39 & 14.47 & 8.07 & 9.88 & 10.27 & \textbf{0.92} \\
WaveNet & \textbf{188.73} & \textbf{13.74} & \textbf{7.8} & \textbf{9.88} & \textbf{10.15} & \textbf{0.92} \\
TFT & 449.58 & 21.2 & 10.98 & 12.61 & 13.01 & 0.82 \\
TST & 1031.83 & 32.12 & 19 & 20.44 & 21.5 & 0.58 \\
\hline
\end{tabular}
\end{minipage} &
\begin{minipage}{0.48\linewidth}
\centering
\caption*{\textbf{(b) Medical Clinic}}
\vspace{-5pt}
\begin{tabular}{l c c c c c c}
\hline
\textbf{Model} & \textbf{MSE} & \textbf{RMSE} & \textbf{MAE} & \textbf{SMAPE} & \textbf{MAPE} & \textbf{R\textsuperscript{2}} \\
\hline
LSTM & 9.56 & 3.09 & 2.28 & 12.04 & 11.95 & 0.87 \\
GRU & 9.04 & 3.01 & 2.25 & 11.83 & 12.08 & 0.88 \\
BLSTM & 8.1 & 2.85 & 2.15 & 11.34 & 11.5 & 0.89 \\
BGRU & 8.64 & 2.94 & 2.14 & 11.15 & 11.2 & 0.89 \\
CNN & 12.09 & 3.48 & 2.51 & 12.36 & 11.65 & 0.84 \\
TCN & \textbf{5.19} & \textbf{2.28} & \textbf{1.62} & \textbf{8.48} & \textbf{8.46} & \textbf{0.93} \\
DCNN & 7.23 & 2.69 & 1.97 & 10.19 & 9.87 & 0.91 \\
WaveNet & 5.26 & 2.29 & 1.63 & 8.75 & 8.76 & \textbf{0.93} \\
TFT & 23.67 & 4.87 & 3.39 & 17.39 & 16.63 & 0.69 \\
TST & 9.63 & 3.1 & 2.3 & 11.98 & 12.36 & 0.87 \\

\hline
\end{tabular}
\end{minipage} \\
\\
\vspace{20pt}

\noindent \begin{minipage}{0.48\linewidth}
\centering
\caption*{\textbf{(c) Retail Store}}
\vspace{-5pt}
\begin{tabular}{l c c c c c c}
\hline
\textbf{Model} & \textbf{MSE} & \textbf{RMSE} & \textbf{MAE} & \textbf{SMAPE} & \textbf{MAPE} & \textbf{R\textsuperscript{2}} \\
\hline
LSTM & 0.49 & 0.7 & 0.45 & 15.45 & 16.45 & 0.84 \\
GRU & 0.53 & 0.73 & 0.5 & 17.33 & 18.76 & 0.83 \\
BLSTM & 0.55 & 0.74 & 0.47 & 15.8 & 16.46 & 0.82 \\
BGRU & 0.49 & 0.7 & 0.47 & 16.41 & 16.97 & 0.84 \\
CNN & 0.42 & 0.65 & 0.44 & 15.51 & 16.85 & \textbf{0.86} \\
TCN & 0.43 & 0.66 & \textbf{0.43} & 15.15 & \textbf{15.18} & \textbf{0.86} \\
DCNN & \textbf{0.41} & \textbf{0.64} & \textbf{0.43} & 15.44 & 16.24 & \textbf{0.86} \\
WaveNet & 0.42 & 0.65 & \textbf{0.43} & 15.6 & 15.96 & \textbf{0.86} \\
TFT & 1.04 & 1.02 & 0.66 & 21.7 & 24.46 & 0.66 \\
TST & 0.47 & 0.69 & 0.44 & \textbf{14.89} & 15.96 & 0.84 \\

\hline
\end{tabular}
\end{minipage} &
\begin{minipage}{0.48\linewidth}
\centering
\caption*{\textbf{(d) Office Building}}
\vspace{-5pt}
\begin{tabular}{l c c c c c c}
\hline
\textbf{Model} & \textbf{MSE} & \textbf{RMSE} & \textbf{MAE} & \textbf{SMAPE} & \textbf{MAPE} & \textbf{R\textsuperscript{2}} \\
\hline
LSTM & 44.42 & 6.66 & 4.81 & 11.5 & 12.11 & 0.86 \\
GRU & 40.8 & 6.39 & 4.65 & 11.17 & 11.89 & 0.87 \\
BLSTM & 48.7 & 6.98 & 5.13 & 12.68 & 13.75 & 0.85 \\
BGRU & 38.19 & 6.18 & 4.5 & 10.94 & 11.36 & 0.88 \\
CNN & 23.33 & 4.83 & 3.59 & 8.69 & 8.94 & 0.93 \\
TCN & 20.99 & 4.58 & 3.35 & 8.12 & 8.26 & 0.93 \\
DCNN & 26.19 & 5.12 & 3.8 & 9.4 & 9.49 & 0.92 \\
WaveNet & \textbf{19.84} & \textbf{4.45} & \textbf{3.22} & \textbf{7.88} & \textbf{8.05} & \textbf{0.94} \\
TFT & 72.87 & 8.54 & 6.16 & 14.2 & 14.83 & 0.77 \\
TST & 95.46 & 9.77 & 6.97 & 15.62 & 17 & 0.7 \\
\hline
\end{tabular}
\end{minipage}

\end{tabular}
\label{tab:accuracyOn4Datasets}
\end{table*}

To examine the performance of the anomaly detection models on the remaining four datasets, a manufacturing facility, a medical clinic, a retail store, and an office building, Table \ref{tab:accuracyOn4Datasets} presents the key performance metrics for each model type. Note that metrics such as MSE, RMSE, and MAE should not be compared among datasets and those are scale-dependent metrics. No single algorithm outperforms all others across all performance metrics and datasets; nevertheless, TCN achieves overall excellent performance. We evaluate our expandability approach across all listed models.

\subsection{Global Feature Importance} \label{sec:result_globalfeatureimportance}

\begin{figure*}[b!]
    \centering
    \includegraphics[width=1\linewidth]{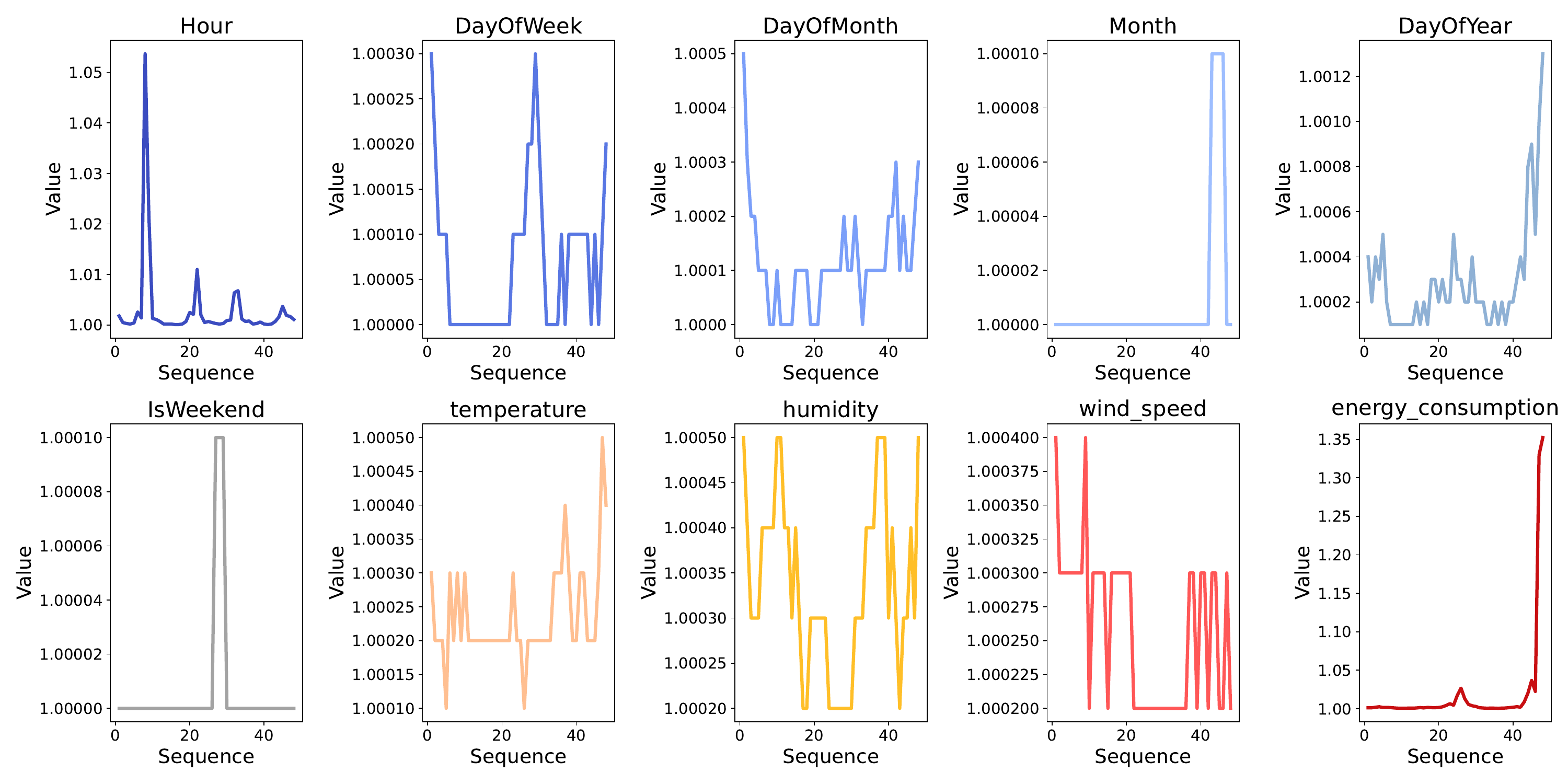}
    
        \captionsetup{skip=4pt} 
    \caption{Global Feature Importance Line Plot of Sequence Windows for LSTM Model on residential dataset}
    \label{fig:Global_Feature_Figure}
\end{figure*}

In this section, we present the global feature importance results for 10 features across 48 time sequences, offering a comprehensive view of each feature’s contribution to model prediction. As detailed in the methodology, we applied an exponential transformation to the Random Forest results to better highlight the distinctions in feature importance. Based on Figure \ref{fig:Global_Feature_Figure} for the residential data, the previous hours energy consumption data emerged as the most significant predictor for the LSTM model. The energy consumption feature exhibited a consistently increasing trend in importance over the last 48 hours, peaking at values of 1.352 and 1.329 for the final time steps, highlighting its relevance near the prediction window (e.g., sequence of 48 hours before prediction). The hour feature also exhibited notable influence, with peak values of 1.0068 and 1.0064. Other features, while less influential, follow consistent temporal trends. These observations highlight the importance of key features in selecting a contextually relevant baseline for explainability.
This process was repeated for other deep learning models and other datasets, consistently revealing that energy consumption history and time-related features are critical for predictions due to the inherent temporal patterns in energy consumption. These global feature importance results will guide the following sections, where we focus on selecting a background dataset that aligns with the most important features identified in the global model. This selection will enhance the explainability process by concentrating on the features that play a key role in explaining anomalies.

\vspace{-3pt}

\subsection{Explanation Step: Comparing Random Background Dataset and Similar Background Dataset Selection}


In this section, we provide a detailed comparison between the random and similar background dataset selection methods by analyzing a specific anomaly point identified using the LSTM model on the residential data, as shown in Figures \ref{fig:LSTM_General} and \ref{fig:LSTM_Similar}.

As discussed in Section \ref{sec:result_globalfeatureimportance}, these feature importance have dimensions of 48$\times$10, where the features are represented in the rows, and their sequences are displayed along the columns. The features are sorted by the absolute value of their SHAP values, with the most impactful features positioned at the top.
For example, in Figure \ref{fig:LSTM_General}, the date and time features, such as the month, are identified as the most important in explaining the anomaly. In contrast, in Figure \ref{fig:LSTM_Similar}, energy consumption and weather-related features are more significant contributors. This demonstrates that the similar background dataset selection can capture different aspects of the data, such as weather information, due to its temporal alignment.

Another important aspect of these heatmaps is their relevance to the reliability of the results. Before explaining this further, it is important to clarify the role of the line plots and the dotted horizontal line present at the top of each heatmap.
As outlined in Equation \ref{eq:prediction}, the prediction function $f(x)$ (represented by the line plot) is the sum of the SHAP values, showing how each feature shifts the prediction either away from or towards the base value (dotted horizontal line). Given that the dataset consists of 10 features and 48 time steps, the prediction function is defined as:

\begin{equation} \label{eq:time_series_decomposition}
f(x) = \mathrm{Base Value} + \sum_{t=1}^{48} \sum_{i=1}^{10} \mathrm{SHAP}_t(i)
\end{equation}

Here, the \textit{Base Value} represents the average prediction derived from the background dataset, and \( \mathrm{SHAP}_t(i) \) represents the contribution of feature \( i \) at time step \( t \). This breakdown provides a clear view of how each feature at each time step influences the overall prediction.

From the heatmaps in Figures \ref{fig:LSTM_General} and \ref{fig:LSTM_Similar}, which correspond to the same anomaly detected by the LSTM model, the prediction, true value, and error are 1.60, 4.75, and -3.15, respectively. Since the model’s prediction is lower than the true value, positive SHAP values (red cells) represent features that adjust the prediction closer to the true value (offsets), while negative SHAP values (blue cells) correspond to features that push the prediction further from the true value, contributing to the anomaly.

In the following, we compare the random and similar background dataset selection based on these aspects:



\begin{figure*}
    \includegraphics[width=0.95\linewidth]{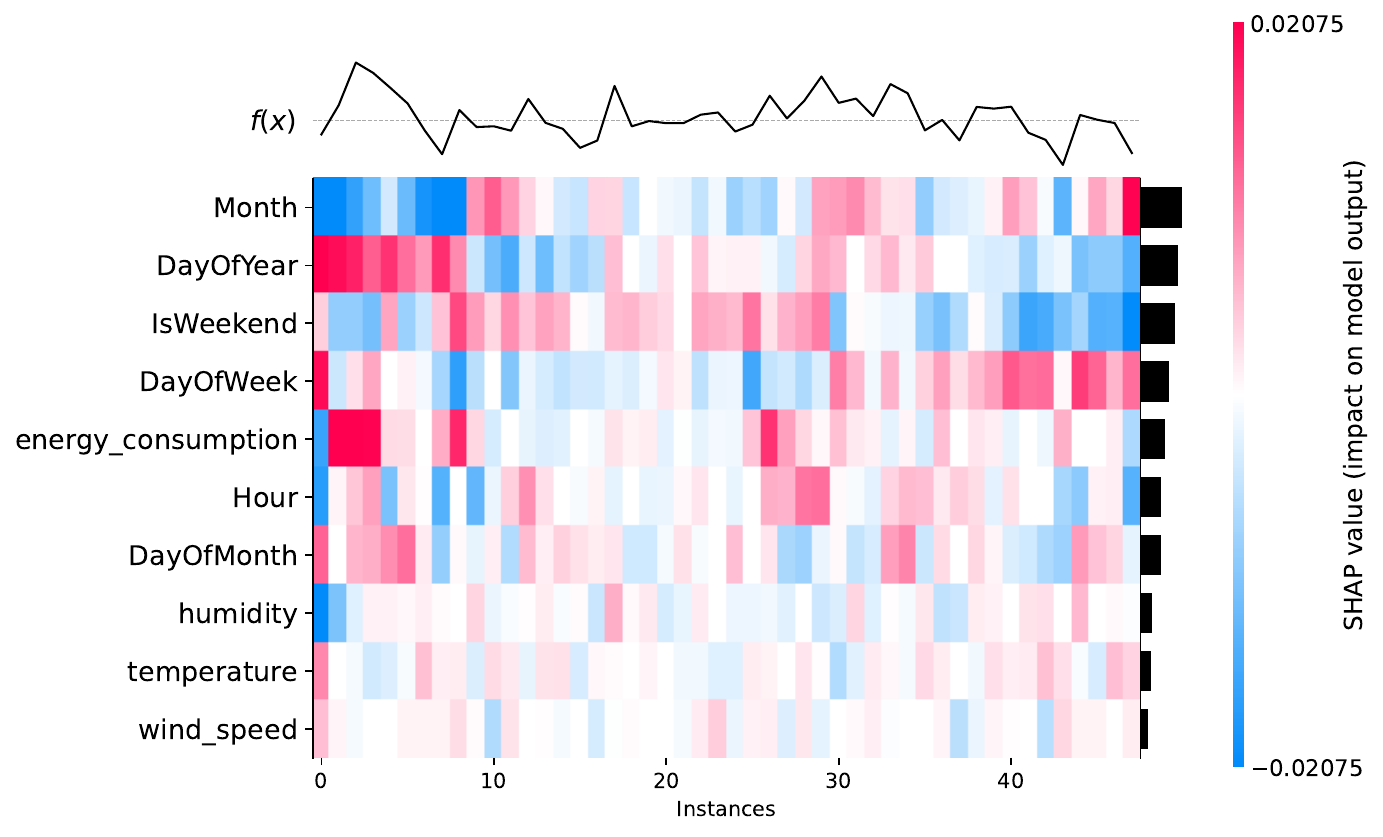}
        \captionsetup{skip=0pt} 
    \caption{SHAP value heatmap for random background dataset selection approach for LSTM Model on residential dataset}.
    \label{fig:LSTM_General}
\end{figure*}

\begin{figure*}
    \centering
    \includegraphics[width=0.95\linewidth]{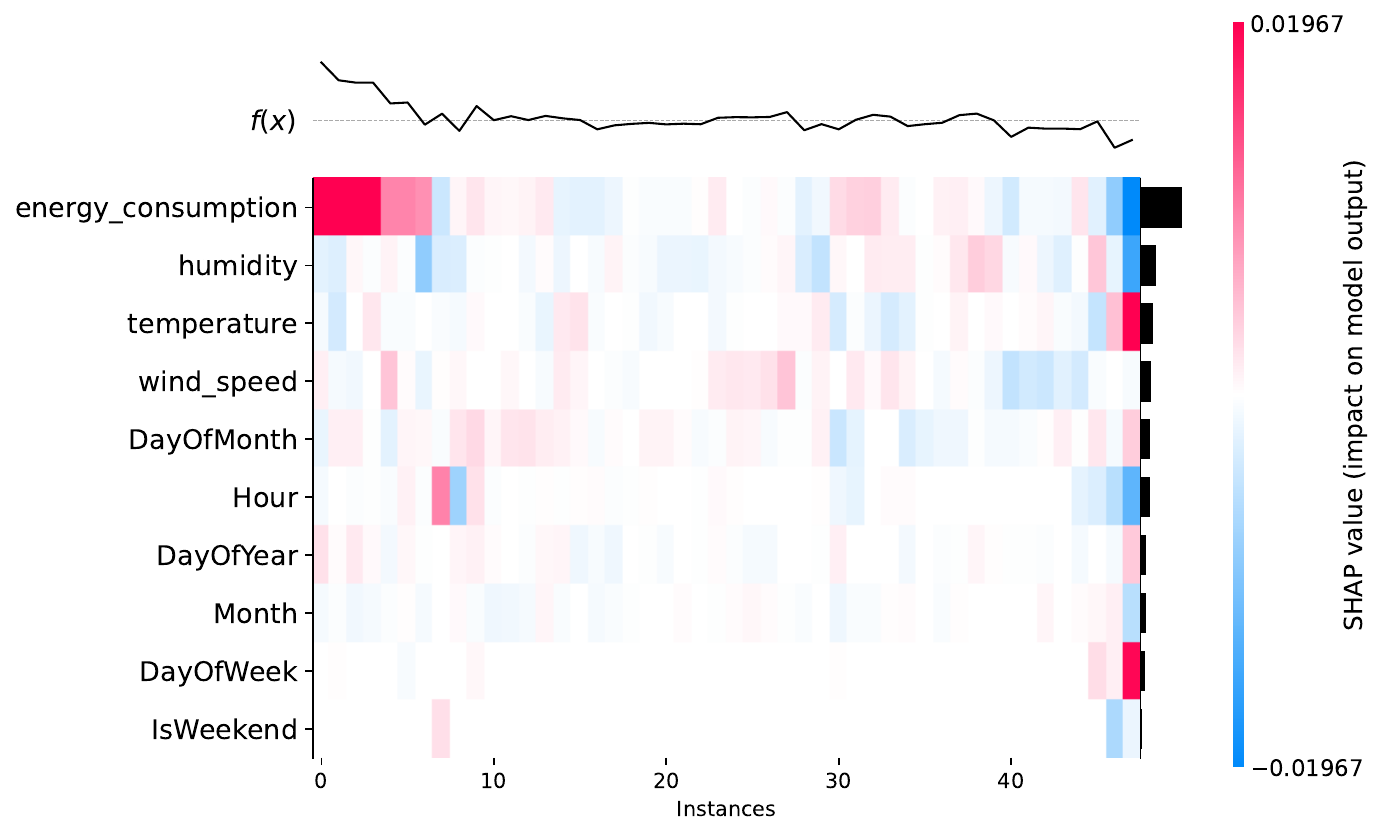}
        \captionsetup{skip=0pt} 
    \caption{SHAP value heatmap for similar background dataset selection approach for LSTM Model on residential dataset}.
    \label{fig:LSTM_Similar}
\end{figure*}


\begin{itemize}
    \item \textbf{Random Background Dataset (Figure \ref{fig:LSTM_General}):} This heatmap shows significant variability in the SHAP values across the sequence of features, with many features showing large positive (bright red) or negative (bright blue) impacts. This variability causes fluctuations in the prediction function $f(x)$ around the base value (dotted line), making it difficult to clearly identify the most important features. The high variability suggests that the randomly selected background dataset may not align closely with the anomaly point, potentially overestimating the importance of less relevant features, such as date and time attributes.
    
\item \textbf{Similar Background Dataset (Figure \ref{fig:LSTM_Similar}):} In contrast, the heatmap from the similar background dataset shows a more stable pattern (fewer fluctuations of $f(x)$ around the base value), with energy consumption and weather-related features emerging as the primary drivers of the model's predictions. This approach reduces the noise caused by less relevant features (more features near zero as most of them with light red or blue colors), offering a clearer identification of the factors contributing to the anomaly. For instance, energy consumption functions as an offset, correcting the model’s prediction towards the true value, while weather features such as humidity, temperature, and wind speed reflect genuine variations that contribute to the anomaly.
\end{itemize}


These observations underscore the effectiveness of using similar background dataset selected according to the proposed similarity metrics in enhancing the interpretability of SHAP values. By focusing on a relevant background dataset, we can better isolate the features that genuinely influence the model's predictions, reducing the impact of unrelated features. This approach is especially valuable given the complexity of the dataset, which includes 480 features (48 sequences $\times$ 10 features). Using appropriate background dataset allows for more focused and meaningful explanations of the model's behavior.

\subsection{SHAP Density Plot Analysis}

\begin{figure*}[b!]
    \centering
    \includegraphics[width=0.99\linewidth]{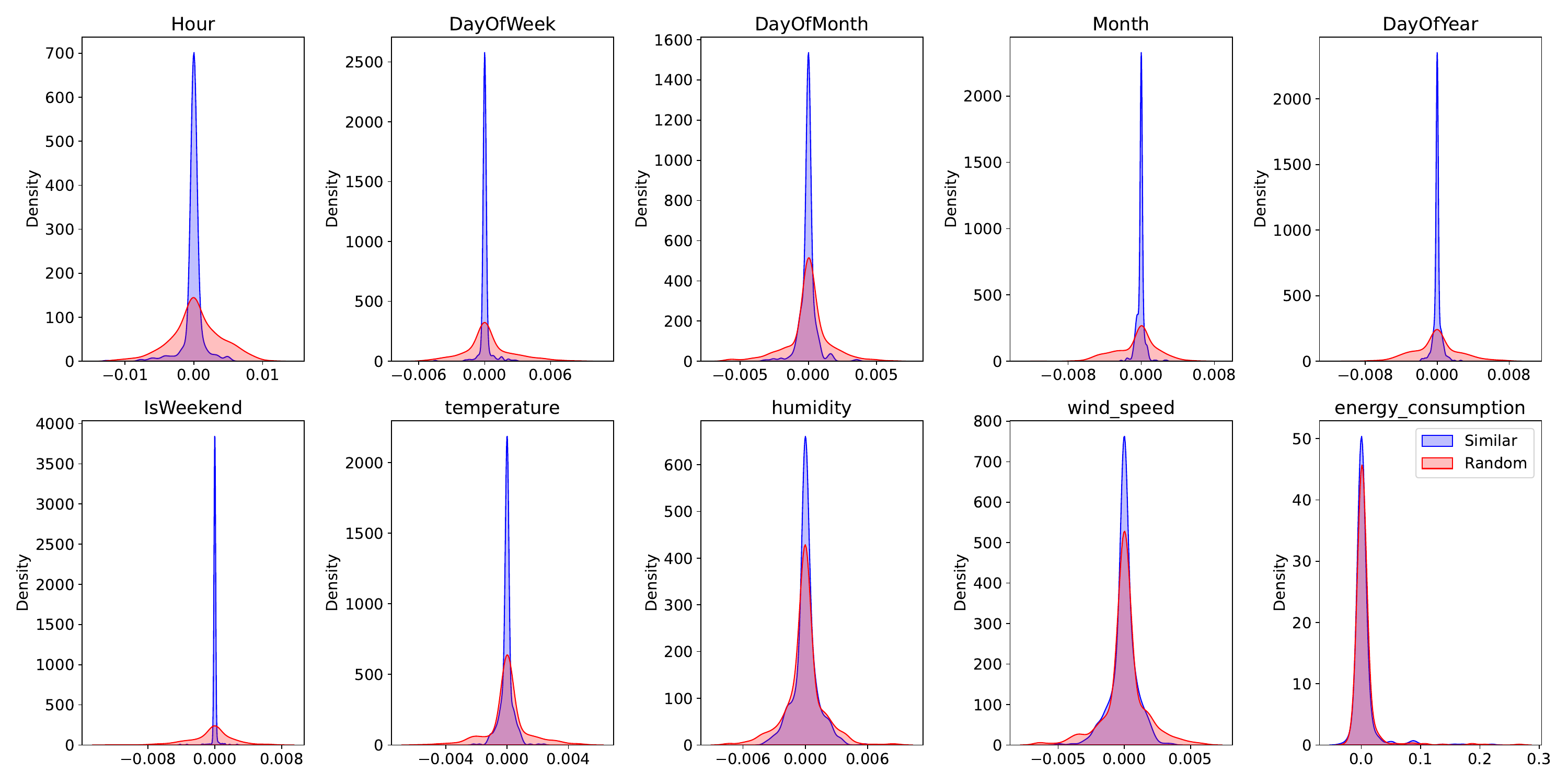}
        \captionsetup{skip=0pt} 

    \caption{Density plots of SHAP values for similar vs. random background dataset across 10 features, based on anomalies detected by the LSTM model on residential dataset}.
    \label{fig:shap_density_plots}
\end{figure*}

As seen from the density plots in Figure \ref{fig:shap_density_plots}, based on anomalies detected by the LSTM model on residential dataset, a notable reduction in the variation of SHAP values is observed for date and time-related features (Hour, DayOfWeek, DayOfMonth, Month, DayOfYear) when utilizing similar background dataset. This reduction underscores the ability of this approach to filter out less relevant features, thereby allowing the model to focus on factors more closely linked to the detected anomalies, such as weather information. However, while the variation is significantly reduced, these features are not entirely disregarded. In cases where energy consumption is unusual at specific times, these features still contribute to the model’s prediction, albeit with less fluctuation.


This behavior indicates that the similar background dataset essentially serves as a filter, helping to emphasize the most impactful features while minimizing noise from less relevant ones. This filtering effect is particularly beneficial when working with high-dimensional data, where many features may not be directly related to the anomalies. By concentrating the SHAP values near zero for these date and time features, the model can more effectively highlight the key drivers of unusual energy consumption, such as weather conditions or specific patterns in energy use.

\subsection{Stability of SHAP Explanations across Multiple Datasets and Explainability Approaches}



Analysis of results so far focused on a residential dataset, demonstrating the benefit of our technique for achieving robust explanations. Building on this, here we expand the analysis to the remaining four datasets-- a manufacturing facility, a medical clinic, a retail store, and an office building--to examine the portability of our technique across diverse energy consumer types. Additionally, for each dataset, we consider diverse XAI algorithms to illustrate the wide applicability of our technique.

Moreover, to further evaluate the impact of our approach for background dataset selection methods on SHAP values, we conducted a comparison between random and similar background dataset approaches across different deep learning models and datasets from diverse energy consumer types. This analysis explored various explainability algorithms, including SHAP variants (Kernel, Partition, and Sampling), as well as LIME and Permutation, applied across all five datasets.

Table \ref{tab:variance_reduction} presents the results. For each dataset and each XAI method, the table reports the mean and standard deviation (Mean ± SD) of SHAP/LIME value variability for random baseline and our baseline, calculated as the standard deviation of feature importance across detected anomalies. For all datasets and all XAI methods, our technique achieves lower variability than the random baseline, highlighting its ability to provide more stable and reliable explanations compared to the random baseline. Additionally, the table includes the percentage reduction in variability, calculated as the relative decrease in the similar baseline's mean compared to the random baseline, to quantify the effectiveness of the proposed method. Bartlett’s test results (P-Value) \cite{bartlett1937properties} are also presented to assess the statistical significance of the differences in variability between the two approaches, random baseline and our baseline. A single * indicates significance on the 5\% level and ** denotes significance on the 1\% level. 

For the residential dataset, significant reductions in variability were observed with Kernel (44\%, $P = 0.009$), Permutation (46.7\%, $P = 0.029$), and Sampling SHAP (46.7\%, $P = 0.001$), while LIME (48.1\%, $P = 0.119$) and Partition (26.1\%, $P = 0.138$) showed notable reductions without statistical significance. In the manufacturing facility dataset, the proposed method achieved particularly strong results, with Partition (80.3\%, $P < 0.001$), Kernel (70.3\%, $P < 0.001$), Permutation (56.8\%, $P < 0.001$), and Sampling SHAP (57.6\%, $P < 0.001$) all demonstrating significant variability reductions. LIME, although less impactful, still achieved a reduction of 21.1\% ($P = 0.243$).

The medical clinic dataset showed significant improvements for Kernel (35.7\%, $P = 0.028$), Permutation (36.2\%, $P = 0.039$), and Sampling SHAP (36.5\%, $P = 0.015$), with LIME (17.9\%, $P = 0.242$) and Partition SHAP (18.6\%, $P = 0.198$) exhibiting modest reductions. The retail store dataset presented mixed results, with statistical significance only achieved for Partition SHAP (33.3\%, $P < 0.001$). Kernel (32.4\%, $P = 0.567$), Sampling (25.8\%, $P = 0.563$), and Permutation SHAP (22.6\%, $P = 0.530$) demonstrated modest reductions without significance, while LIME showed minimal impact (4.2\%, $P = 0.453$). Finally, for the office building dataset, significant improvements were observed with Kernel (52.1\%, $P = 0.002$), Permutation (46.7\%, $P < 0.001$), and Sampling SHAP (47.8\%, $P < 0.001$), while LIME (36.1\%, $P = 0.038$) also achieved a statistically significant reduction. Partition SHAP, however, showed limited impact with a reduction of 10.5\% ($P = 0.133$).

Overall, our approach reduced the variability of the explanations across all datasets and across all XAI methods, with most differences being statistically significant. Particularly good improvements were observed for Kernel, Permutation, and Sampling SHAP. For some datasets and XAI methods, the reduction was larger than for others, but it was present for all datasets. These findings underscore the robustness of the baseline selection strategy in reducing variability and enhancing the stability of SHAP explanations across multiple datasets and XAI methods.

\renewcommand{\arraystretch}{1.5}  
\begin{table*}[ht]
    \centering
    \setlength{\tabcolsep}{2pt}
    \caption{\\
    Comparison of variance reduction and statistical results across datasets and XAI methods.}
    \vspace{-5pt}
    {\fontsize{8.55}{8.55}\selectfont  
        \begin{tabular}{llcclc}
            \toprule
            \textbf{Dataset} & 
            \textbf{XAI methods} & 
            \makecell{\textbf{Random Baseline} \\ (mean $\pm$ sd)} & 
            \makecell{\textbf{Proposed Baseline} \\ (mean $\pm$ sd)} & 
            \makecell{\textbf{Statistical Test} \\ (P-Value)} & 
            \makecell{\textbf{Reduction} \\ (\%)} \\
            \midrule
            Residential & Kernel & 0.050 $\pm$ 0.110 & 0.028 $\pm$ 0.034 & 27.851 (0.009*) & 44.0 \\
             & Lime & 0.077 $\pm$ 0.192 & 0.040 $\pm$ 0.075 & 27.773 (0.119) & 48.1 \\
             & Partition & 0.023 $\pm$ 0.064 & 0.017 $\pm$ 0.026 & 219.271 (0.138) & 26.1 \\
             & Permutation & 0.045 $\pm$ 0.105 & 0.024 $\pm$ 0.031 & 27.326 (0.029*) & 46.7 \\
             & Sampling & 0.045 $\pm$ 0.105 & 0.024 $\pm$ 0.031 & 27.646 (0.001**) & 46.7 \\
            \midrule
            Manufacturing facility & Kernel & 1.214 $\pm$ 1.932 & 0.360 $\pm$ 0.667 & 23.598 (0.000**) & 70.3 \\
             & Lime & 0.478 $\pm$ 1.233 & 0.377 $\pm$ 1.015 & 8.857 (0.243) & 21.1 \\
             & Partition & 0.238 $\pm$ 0.766 & 0.047 $\pm$ 0.113 & 59.502 (0.000**) & 80.3 \\
             & Permutation & 0.759 $\pm$ 2.046 & 0.328 $\pm$ 0.681 & 25.524 (0.000**) & 56.8 \\
             & Sampling & 0.739 $\pm$ 2.037 & 0.313 $\pm$ 0.667 & 24.407 (0.000**) & 57.6 \\
            \midrule
            Medical clinic & Kernel & 0.126 $\pm$ 0.267 & 0.081 $\pm$ 0.125 & 17.410 (0.028*) & 35.7 \\
             & Lime & 0.117 $\pm$ 0.290 & 0.096 $\pm$ 0.226 & 2.562 (0.242) & 17.9 \\
             & Partition & 0.043 $\pm$ 0.134 & 0.035 $\pm$ 0.084 & 12.215 (0.198) & 18.6 \\
             & Permutation & 0.116 $\pm$ 0.264 & 0.074 $\pm$ 0.121 & 17.969 (0.039*) & 36.2 \\
             & Sampling & 0.115 $\pm$ 0.262 & 0.073 $\pm$ 0.121 & 17.622 (0.015*) & 36.5 \\
            \midrule
            Retail store & Kernel & 0.034 $\pm$ 0.046 & 0.023 $\pm$ 0.036 & 2.662 (0.567) & 32.4 \\
             & Lime & 0.024 $\pm$ 0.037 & 0.023 $\pm$ 0.030 & 2.391 (0.453) & 4.2 \\
             & Partition & 0.015 $\pm$ 0.014 & 0.010 $\pm$ 0.000 & 757.685 (0.000**) & 33.3 \\
             & Permutation & 0.031 $\pm$ 0.049 & 0.024 $\pm$ 0.036 & 2.938 (0.530) & 22.6 \\
             & Sampling & 0.031 $\pm$ 0.047 & 0.023 $\pm$ 0.036 & 2.856 (0.563) & 25.8 \\
             \midrule
            Office building & Kernel & 0.219 $\pm$ 0.410 & 0.105 $\pm$ 0.182 & 24.249 (0.002**) & 52.1 \\
             & Lime & 0.244 $\pm$ 0.577 & 0.156 $\pm$ 0.306 & 15.670 (0.038*) & 36.1 \\
             & Partition & 0.019 $\pm$ 0.033 & 0.017 $\pm$ 0.024 & 38.760 (0.133) & 10.5 \\
             & Permutation & 0.195 $\pm$ 0.426 & 0.104 $\pm$ 0.183 & 26.349 (0.000**) & 46.7 \\
             & Sampling & 0.201 $\pm$ 0.448 & 0.105 $\pm$ 0.184 & 28.660 (0.000**) & 47.8 \\
            \bottomrule
        \end{tabular}
    }
    \label{tab:variance_reduction}
    \vspace{0.1cm}
    \parbox{\textwidth}{ ** indicates significance at the 1\% level; * indicates significance at the 5\% level.}
\end{table*}
\renewcommand{\arraystretch}{1}  

\section{Conclusion} \label{sec:Conclusion}

This study presents a comprehensive methodology for explaining deep learning-based anomaly detection in energy consumption data. Through a systematic approach that integrates advanced anomaly detection techniques and innovative explanation methods leveraging a novel approach for selecting the background dataset (baseline) for model-agnostic explainability algorithms, the study provides improved explainability of detected anomalies applicable across a diverse range of deep learning models, SHAP variants, and energy consumption datasets. The key innovation lies in the selection of context-relevant information for the baseline employed in explaining anomalies. Moreover, we guide the explanations toward features highly relevant in the deep learning model. This strategy ensures more consistent explanations, enabling a deeper understanding of the factors driving anomaly detection.

The evaluation conducted on five datasets, with five XAI approaches (Kernel SHAP, Partition SHAP, Sampling SHAP, LIME, Permutation) demonstrated that our baseline approach significantly reduces variability in feature attributions, with reductions ranging from 26.1\% to 80.3\%, depending on the dataset and a SHAP variant. The statistical analysis validated that our approach, compared to the random baseline, achieves a significant reduction in variability for Kernel, Permutation, and Sampling SHAP across most datasets. Although LIME and Partition SHAP did not always achieve statistical significance, they still exhibited reductions in variability.

Future work will examine developing additional metrics for evaluating the quality of explanations as well as examining human perception of the evaluations. Moreover, the approach will be examined with different use cases.

\section*{CRediT authorship contribution statement}
\textbf{Mohammad Noorchenarboo}: Conceptualization, Methodology, Data analysis and curation, Software, Writing – original draft and editing. \textbf{Katarina Grolinger}: Conceptualization, Methodology, Supervision, Writing – review and editing, Guidance, and Oversight throughout the research process.


\section*{Acknowledgement}
This work was supported by London Hydro and Natural Sciences and Engineering Research Council of Canada (NSERC) under grant ALLRP 571890 - 21. Computation was enabled in part by the Digital Research Alliance of Canada. 

\section*{Data availability}
The data used in this study is proprietary and subject to confidentiality agreements; therefore, it cannot be shared.

\bibliographystyle{elsarticle-num}
\bibliography{Refs_updated.bib}

\end{document}